\documentclass[twocolumn]{svjour3}         
\usepackage[dvipsnames,usenames]{color}
\smartqed  
\usepackage{graphicx}
 \usepackage{url} 
\usepackage{times}
\usepackage{epsfig} 
\usepackage{amsmath}
\usepackage{amssymb}
\usepackage{float}
\usepackage[percent]{overpic}
\usepackage[linesnumbered,boxed]{algorithm2e}

\usepackage[pagebackref=true,breaklinks=true,colorlinks,bookmarks=false]{hyperref}

\journalname{International Journal of Computer Vision}

\begin{document}

\title{3D Hand Pose Detection in Egocentric RGB-D Images
\thanks{This work was supported by
EU grant Egovision4Health.}  
} 


\author{Gr\'egory Rogez  \and
        J. S. Supan\v{c}i\v{c} III  \and
        Maryam Khademi \and  \\
        J. M. M. Montiel  \and
        Deva Ramanan
}


\institute{Gr\'egory Rogez, J. M. M. Montiel \at
               Aragon Institute of Engineering Research (i3A), Universidad de Zaragoza, Spain \\
	   \email{\{grogez,josemari\}@unizar.es}
           \and
           Gr\'egory Rogez, J. S. Supan\v{c}i\v{c} III, Maryam Khademi, Deva Ramanan \at
              Dept. of Computer Science, University of California, Irvine, USA \\
			  \email{\{grogez,jsupanci,mkhademi,dramanan\}@unizar.es}
              }

\date{Received: date / Accepted: date}

\maketitle

\begin{abstract}
We focus on the task of everyday hand pose estimation from egocentric viewpoints. For this task, we show that depth sensors are particularly informative for extracting near-field interactions of the camera wearer with his/her environment.
Despite the recent advances in full-body pose estimation using Kinect-like sensors,
reliable monocular hand pose estimation in RGB-D images is still an unsolved problem.
The problem is considerably exacerbated when analyzing hands performing daily activities from a first-person viewpoint, due to severe occlusions arising from object manipulations and a limited field-of-view.
Our system addresses these difficulties by exploiting strong priors over viewpoint and pose in a discriminative tracking-by-detection framework. Our priors are operationalized through a photorealistic synthetic model of egocentric scenes, which is used to generate training data for learning depth-based pose classifiers.
%
We evaluate our approach on an annotated dataset of real egocentric object manipulation scenes and compare to both commercial and academic
approaches. Our method provides state-of-the-art performance for both hand detection and pose estimation in egocentric RGB-D images.
 


\end{abstract}

\section{Introduction}
Much recent work has explored various applications of egocentric RGB cameras, spurred on in part by the availability of low-cost mobile sensors such as Google Glass, Microsoft SenseCam, and the GoPro camera. Many of these applications, such as life-logging \cite{hodges2006sensecam}, medical rehabilitation~\cite{YangSL10}, and augmented reality~\cite{BerghG11}, require inferring the interactions of the first-person observer with his/her environment while recognizing his/her activities. Whereas third-person-view activity analysis is often driven by human full-body pose, egocentric activities are often defined by hand pose and the objects that the camera wearer interacts with.  Towards that end, we specifically focus on the tasks of hand detection and hand pose estimation from egocentric viewpoints of daily activities. We show that depth-based cues, extracted from an {\em egocentric depth camera}, provides an extraordinarily helpful cue for egocentric hand-pose estimation.
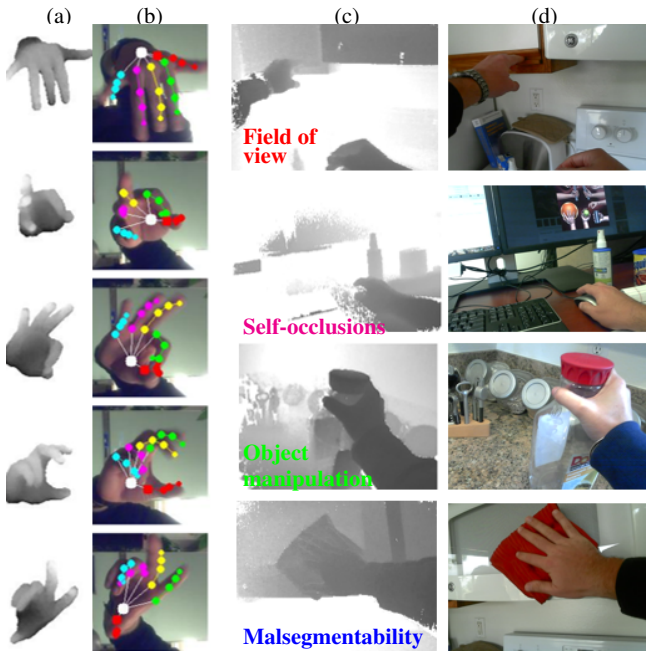
\begin{figure}[htb]
  \centering 
 \vspace{-5pt}\hspace{-20pt} (a) \hspace{20pt}  (b) \hspace{60pt} (c)  \hspace{60pt} (d) \hspace{50pt}
 \begin{overpic}[width=\columnwidth]{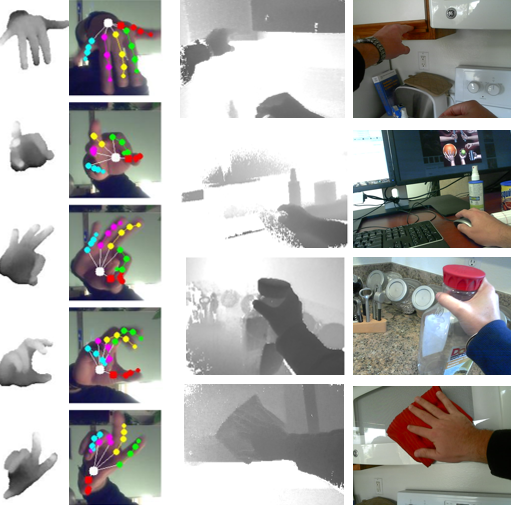}
    \put(37,1){\color{blue} \bf Malsegmentability }
     \put(37,30){\color{green} \bf Object }
   \put(37,26){\color{green} \bf  manipulation }
   \put(37,50){\color{magenta} \bf Self-occlusions } 
   \put(37,80){\color{red} \bf Field of } 
   \put(37,77){\color{red} \bf view } 
  \end{overpic} 
  \caption{\footnotesize {\bf Challenges}. We contrast third person depth (a) and RGB images (b) (overlaid with pose estimates from \cite{Qian0WT014}) with depth image and RGB images (c,d) from egocentric views of daily activities. {\color{red} Hands leaving the field-of-view}, {\color{magenta} self-occlusions}, {\color{green} occlusions due to objects } and {\color{blue} malsegmentability due to interactions with the environment} are common hard cases in egocentric settings.}
    \label{fig:challenge}
\end{figure}



One may hope that depth simply ``solves'' the problem, based on successful systems for real-time human pose estimation based on Kinect sensor~\cite{ShottonFCSFMKB11} and prior work on articulated hand pose estimation for RGB-D sensors \cite{bmvc2011oikonom,TangYK13,KeskinKKA_ECCV12,Qian0WT014,TangYK13}. Recent approaches have also tried to exploit the 2.5D data from Kinect-like devices to understand complex scenarios such as
object manipulation \cite{KyriazisA13} or  two interacting hands \cite{Oikonomidis2012}. We show that various assumptions about visibility/occlusion and manual tracker initialization may not hold in an egocentric setting, making the problem still quite challenging.

{\bf Challenges:} Three primary challenges arise for hand pose estimation in everyday egocentric views, compared to 3rd person views. First, {\em tracking is less reliable}. Even assuming that manual initialization is possible, a limited field-of-view from an egocentric viewpoint causes hands to frequently move outside the camera view frustum. This makes it difficult to apply tracking models that rely on accurate estimates from previous frames, since the hand may not even be visible. Second, {\em active hands are difficult to segment.} Many previous systems for 3rd-person views make use of simple depth-based heuristics to both detect and segment the hand. These are difficult to apply during frames where users interact with objects and surfaces in their environment. Finally, {\em fingers are often occluded} by the hand (and other objects being manipulated) in egocentric views, considerably complicating articulated pose estimation. See examples in Fig.~\ref{fig:challenge}. 

\begin{figure*}[htb]
  \centering
 \hspace{-0mm}\includegraphics[width=\textwidth]{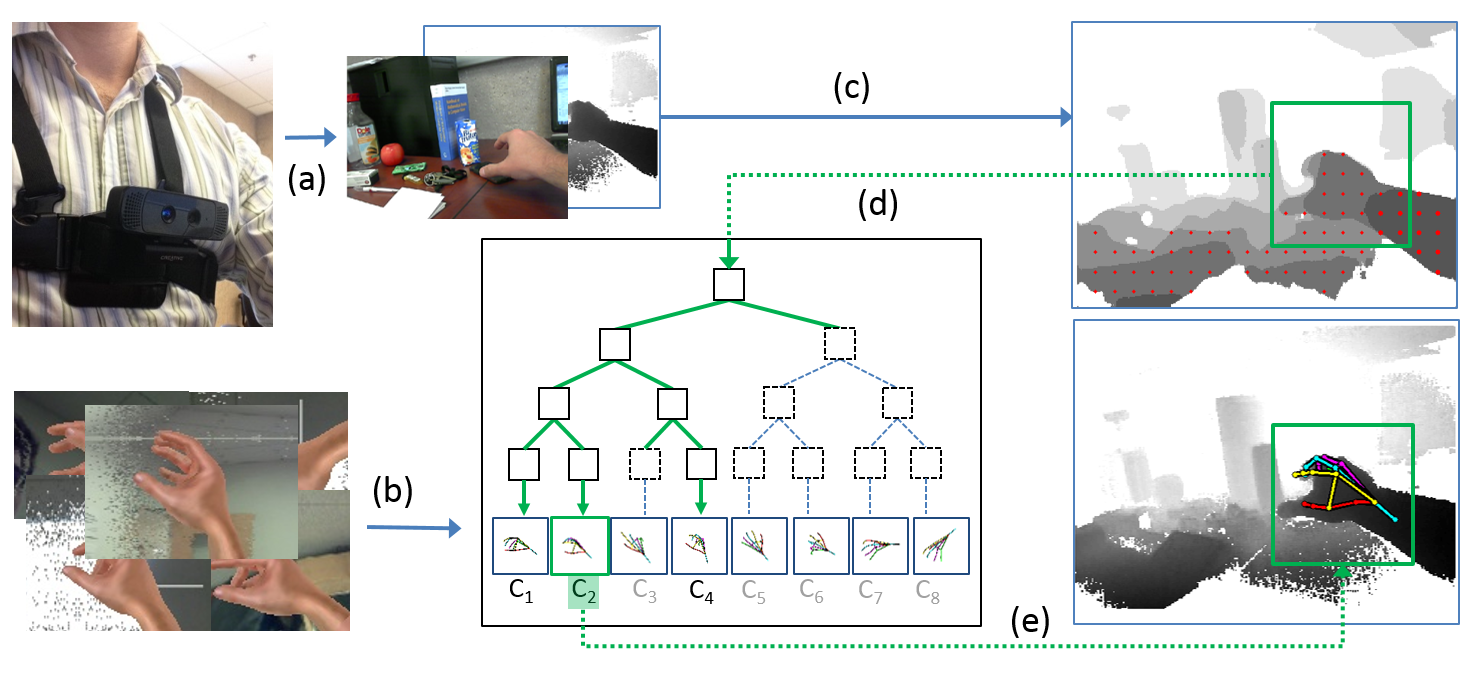}
  \caption{\footnotesize {\bf System overview}. (a) Chest-mounted RGB-D camera. (b) Synthetic egocentric hand
  exemplars are used to define a set of hand pose classes and train a multi-class hand classifier. The depth map is processed to select a sparse
  set of image locations (c) which are classified obtaining a list of probable hand poses (d). Our system produces a final estimate by reporting one or more top-scoring pose classes (e).}
    \label{fig:system}
\end{figure*}
\begin{figure*}[htb]
  \centering
 \hspace{-0mm}\includegraphics[width=\textwidth]{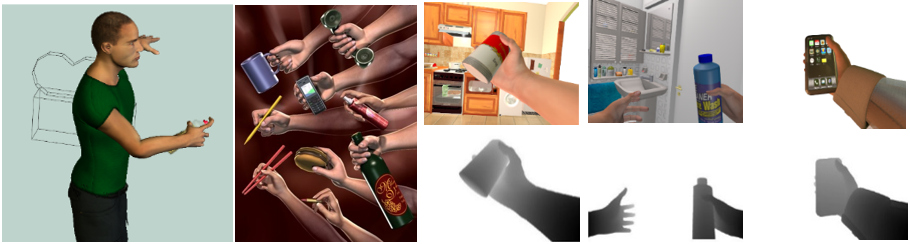} 
 Virtual egocentric camera \hspace{20pt} Everyday hand  package \quad\quad \hspace{50pt} Synthetic egocentric
RGB-D images \quad\quad\quad\quad\quad\quad
  \caption{\footnotesize {\bf Training data}. We show on the left hand side our avatar that we mount with a virtual egocentric camera. In the middle we show the EveryDayHands animation library~\cite{everyhands} used to generate realistic hand-object configurations. On the right, we present some examples  of resulting training images rendered using Poser.}
    \label{fig:trainingdataset}
\end{figure*}

{\bf Our approach:} 
We describe a successful approach to hand-pose
estimation that makes use of the following key observations. First, {\bf depth cues} provide an extraordinarily helpful signal for pose
estimation in the near-field, first-person viewpoints. 
Though this observation may see obvious, state-of-the-art methods for egocentric hand detection do not make use of depth \cite{LiCVPR13,LiICCV13}. Moreoever, in our scenario, depth cues are not ``cheating'' as humans themselves make use of stereopsis for near-field analysis \cite{sakata1999neural,fielder1996does}. Second, the egocentric setting provides strong priors over
{\bf viewpoint, grasps, and interacting-objects}. We operationalize
these priors by generating synthetic training data with a rendered 3D hand model. In contrast to previous work that uses a ``floating hand'', we mount a synthetic egocentric camera to a virtual full-body character interacting with a library of everyday objects. This allows us to make use of contextual cues for both data generation and recognition (see Fig.~\ref{fig:trainingdataset}).
 Third, 
we treat pose estimation (and detection) as a discriminative multi-class classification problem. To efficiently evaluate a large number of pose-specific classifiers, we make use of {\bf hierarchical cascade} architectures.
Unlike much past work, we classify global poses rather than local parts, which allows us to
better reason about self-occlusions. Our classifiers process
single frames, using a tracking-by-detection framework that avoids the need for
manual initialization (see Fig.~\ref{fig:system}c-e).

{\bf Evaluation:} Unlike human pose estimation, there exists no
standard benchmarks for hand pose estimation, especially in egocentric videos. We believe that quantifiable
performance is important for many broader applications such as health-care rehabilitation,
for example. Thus, for the evaluation of our approach, we have
collected and annotated (full 3D hand poses) our own benchmark dataset
of real egocentric object manipulation scenes, which  we will release
to spur further research. It is surprisingly difficult to collect
annotated datasets of hands performing real-world interactions; indeed, many prior work on hand pose estimation evaluate results on synthetically-generated data. We
developed a semi-automatic labelling tool which allows to accurately
annotate partially occluded hands and fingers in 3D, given real-world
RGB-D data. We compare to both commercial and academic
approaches to hand pose estimation, and demonstrate that our method provides state-of-the-art performance for both hand detection and pose estimation in egocentric RGB-D images.

{\bf Overview:} This work is an extension of \cite{RogezKSMR2014}. This manuscript explains our approach with considerably more detail, reviews a broader collection of related work, and provide new and extensive comparisons with state-of-the-art methods. We review related work in Sec.~\ref{sec:work} and present our approach in Sec.~\ref{sec:approach}, focusing on our synthetic training data generation procedure (Sec.~\ref{sec:synth}) and our hierarchical multi-class architecture (Sec.~\ref{sec:hier}). We conclude with experimental results in Sec.~\ref{sec:exp}.

\section{Related work}
\label{sec:work}
\hspace{15pt}{\bf Egocentric hands:} 
Previous work examined the problem of recognizing objects~\cite{fathi2011learning,adl_cvpr12} and interpreting American Sign Language poses \cite{Starner98visualcontextual} from wearable cameras. Much work has also focused on hand detection \cite{LiCVPR13,LiICCV13}, hand tracking \cite{KurataKKJE02,KolschT05,Kolsch10,MorerioMR13}, finger tracking~\cite{DominguezKS06}, and hand-eye tracking \cite{RyooM13} from wearable cameras. Often, hand pose estimation is examined during active object manipulations \cite{MayolBMVC2004,ren2009egocentric,RenG10,fathiunderstanding}. 
Most such previous work makes use of RGB sensors. Our approach demonstrates that an egocentric depth camera makes things considerably easier.

{\bf Egocentric depth:} Depth-based wearable cameras are attractive because depth cues
can be used to better reason about occlusions arising from egocentric
viewpoints. There has been surprisingly little prior work in this vein,
with notable exceptions focusing on targeted applications such as
navigation for the blind~\cite{MannHJLRCD11} and recent work on egocentric object understanding \cite{DamenGMC12,LinHM2014}. We posit that one limitation may be the need for small form-factors for wearable technology, while structured light sensors such as the Kinect often make use of large
baselines. We show that time-of-flight depth cameras are an attractive
alternative for wearable depth-sensing, since they do not require
large baselines and so require smaller form-factors.

{\bf Depth-based pose:} Our approach is closely inspired by the Kinect system and its variants~\cite{ShottonFCSFMKB11}, which makes use of synthetically generated depth maps for articulated pose estimation. Notably, Kinect follows in the tradition of local part models \cite{chenhierarchical,ErolBNBT07}, which are attractive in that they require less training data to model a large collection of target poses. However, it is unclear if local methods can deal with large occlusions (such as those encountered during egocentric object manipulations) where local information can be ambiguous. 
Our approach differs in that our classifiers classify global poses rather than local parts. 
Finally, much previous work assumes that hands are easily segmented or detected. Such assumptions simply do not hold for everyday egocentric interactions.

{\bf Interacting objects:} Estimating the pose of a hand manipulating an object is challenging~\cite{HamerSKG09} due to occlusions and ambiguities in segmenting the object versus the hand. It is attractive to exploit contextual cues through simultaneously tracking hands \cite{KyriazisA13,KyriazisA14} and the object. \cite{BallanTGGP12} use multi-cameras to reduce the number of full occlusions. We jointly model hand and objects with synthetic hand-object exemplars as in \cite{RomeroKEK13}. However, instead of modeling floating hands, we model them in a realistic egocentric context that is constrained by the full human body.


{\bf Tracking vs detection:} Temporal reasoning is also particularly attractive because one can use dynamics to resolve ambiguities arising from self and object occlusions. Much prior work on hand-pose estimation takes this route~\cite{bmvc2011oikonom,TangYK13,KeskinKKA_ECCV12,Qian0WT014}. Our approach differs in that we focus on single-image hand pose estimation, which is required to avoid manual (re)initialization. Exceptions include \cite{XuChe_iccv13,TangCTK14,TompsonSLP14}, who also process single images but focus on third-person views. 

{\bf Generative vs discriminative:} Generative model-based approaches have historically been more popular for hand pose estimation \cite{StengerPAMI06}. A detailed 3D model of the hand pose is usually employed for articulated pose tracking \cite{OikonomidisKA11,Oikonomidis2012} and detailed 3D pose estimation \cite{GorceFP11}. Discriminative approaches \cite{TangYK13,KeskinKKA_ECCV12}
for hand pose estimation tend to require large datasets of training
examples, synthetic, realistic or combined \cite{TangYK13}. Learning formalisms include boosted classifier trees~\cite{OngB04} and randomized decision forests\cite{KeskinKKA_ECCV12}, and regression forests \cite{TangYK13,TangCTK14}. Sridhar et al.~\cite{SridharOT13} propose a hybrid approach that combines discriminative part-based pose retrieval with a generative model-based tracker. Our approach uses a computer graphics model to generate training data, which is then used to learn discriminative pose-specific classifiers.

{\bf Hierarchical cascades:} We approach pose estimation as a hierarchical multi-class classification task, a strategy that dates back at least to Gavrila {\em et al}~\cite{gavrila1999real}. Our framework follows a line of work that focuses on efficient implementation through coarse-to-fine hierarchical cascades~\cite{zehnder2008efficient,stenger2007estimating,chenhierarchical,RogezROT12}. Our work differs in its discriminative training and large-scale ensemble averaging over an exponentially-large set of cascades, both of which considerably improve accuracy and speed.

\section{Our method}
\label{sec:approach}
Our method works by using a computer graphics model to generate synthetic training data. We then use this data to train a classifier for pose estimation. We describe each stage in turn.

\subsection{Synthesizing training data}
\label{sec:synth}

We represent a hand pose as a vector of joint angles of a kinematic skeleton $\theta$. We use a hand-specific forward kinematic model to generate a 3D hand mesh given a particular $\theta$.  In addition to hand pose parameters $\theta$, we also need to specify a camera vector $\phi$ that specifies both a viewpoint and position. We experimented with various priors and various rendering packages.

{\bf Floating hands vs  full-body characters:} Much work on hand pose estimation makes use of an isolated ``floating'' hand mesh model to generate synthetic training data. Popular software packages include the open-source \texttt{\small{libhand}} \cite{libhand} and commercial Poser~\cite{poser,shakhnarovich2003fast}. We posit that modeling a full character body, and specifically, the full arm, will provide important contextual cues for hand pose estimation. To generate egocentric data, we mount a synthetic camera on the chest of a virtual full-body character, naturally mimicking our physical data collection process. To generate data corresponding to different body and hand shapes, we make use of Poser's character library.

{\bf Pose prior:} Our hand model consists of 26 joint angles, $\theta \in [0,2\pi]^{26}$. It is difficult to specify priors over such high-dimensional spaces. We take a non-parametric data-driven approach. We first obtain a training set of joint angles $\{\theta_n\}$ from a collection of  
grasping motion capture data \cite{romero2010spatio}. We then augment this core set of poses with synthetic perturbations, making use of rejection sampling to remove invalid poses. Specifically, we first generate proposals by perturbing the $i^{th}$ joint angle of training sample $n$ with Gaussian noise 
\begin{align}
\theta_n[i] \rightarrow \theta_n[i] + \epsilon \quad \text{where} \quad \epsilon \sim N(0,\sigma_i)
\end{align}
The noise variance $\sigma_i$ is obtained by manual tuning on validation data. 
Notably, we also perturb the {\em entire arm} of the full character-body, which generates natural (egocentric) viewpoint variations of hand configurations. Note that we consider smaller perturbations for fingers to keep grasping poses reasonable. We remove those samples that result in poses that are self-intersecting or lie outside the field-of-view.


{\bf Viewpoint prior:} The above pose perturbation procedure for $\theta$ naturally generates realistic egocentric camera viewpoints $\phi$ for our full character models. We also performed some diagnostic experiments with a floating hand model. To specify a viewpoint prior in such cases, we limited the azimuth $\phi_{az}$ to lie between $180\pm 30$ (corresponding to rear viewpoints), elevation $\phi_{el}$ to lie between $-30$ and $10$ (since hands tend to lie below the chest mount), and bank $\phi_{b}$ to lie between $\pm 30$. We obtained these ranges by looking at a variety of collected data (not used for testing). 

\begin{figure*}[t!]
\begin{tabular}{c}  
       \hspace{-2mm}\includegraphics[width=\textwidth]{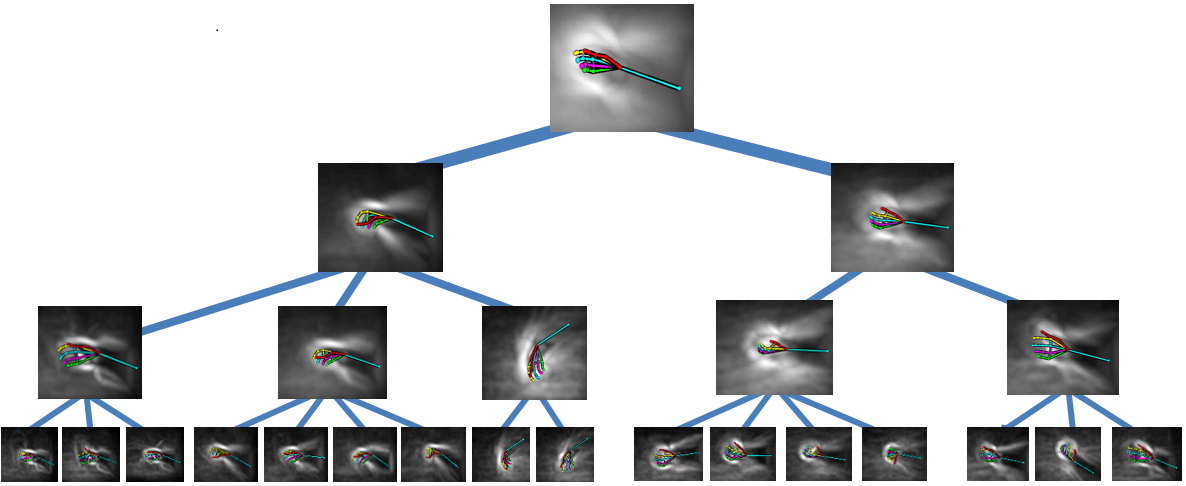}
\end{tabular}
\caption{ {\bf Hierarchy of hand poses}. We visualize a hierarchical graph $G=(V,E)$ of quantized poses with $K=16$ leaves. The $i^{th}$ node in this tree represents a coarse pose class, visualized with the average hand pose and the average gradient map over all the exemplars in that coarse pose.
  \label{fig:hierarchy}
}
\end{figure*} 

{\bf Interacting objects:} We wish to explore egocentric hand pose estimation in the context of natural, functional hand movement. This often involves interactions with the surrounding environment and manipulations of nearby objects. We posit that generating such contextual training data will be important for good test-time accuracy. However, modeling the space of hand grasps and the world of manipulable objects is itself a formidable challenge. We make use of the EveryDayHands animation library~\cite{everyhands}, which contains 40 canonical hand grasps. This package was originally designed as a computer animation tool, but we find the library to cover a reasonable taxonomy of grasps for egocentric recognition. A surprising empirical fact is that humans tend to use a small number of grasps for everyday activities - by some counts, 9 grasps are enough to account for 80\% of human interactions~\cite{ZhengRD11}. Following this observation, we manually amassed a collection of everyday common objects from model repositories \cite{warehouse}. Our objects include spheres and cylinders (of varying sizes), utensils, phones, cups, etc. We paired each object a viable grasp (determined through visual inspection), yielding a final set of 52 hand-object combinations. We apply our rejection-sampling technique to generate a large number of grasp-pose and viewpoint perturbations, yielding a final dataset of 10,000 synthetic egocentric hand-object examples. Some examples are shown in Fig.~\ref{fig:trainingdataset}.

\subsection{Hierarchical cascades}
\label{sec:hier}

We use our training set to learn a model that simultaneously detects hands and estimates their pose. Both tasks are addressed with a scanning window classifier, that outputs one of $K$ discrete pose classes or a background label. One may need a large $K$ to model lots of poses, increasing training/testing times and memory footprints. We address such difficulties through coarse-to-fine sharing and scanning-window cascades. Such architectures have been previously explored in~\cite{zehnder2008efficient,zehnder2008efficient,chenhierarchical,RogezROT12}. We contribute methods for efficient discriminative training, efficient run-time evaluation, and ensemble averaging. To describe our contributions, we first recast previous approaches in a mathematical framework that is amenable to our proposed modifications.

\begin{figure*}[htb!]
\begin{tabular}{cc} 
\hspace{-3mm}\includegraphics[width=0.5\textwidth]{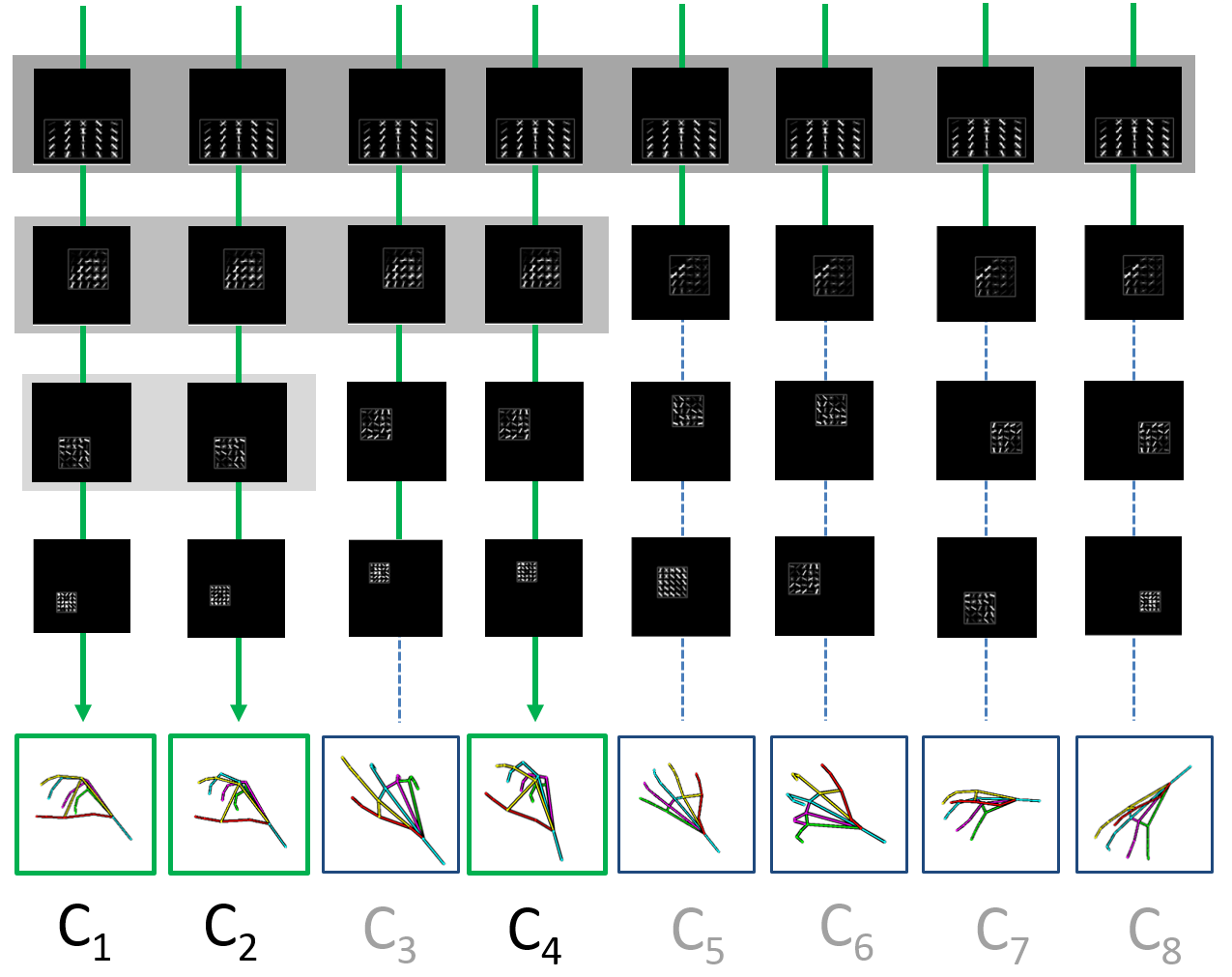}&\hspace{-2mm}\includegraphics[width=0.5\textwidth]{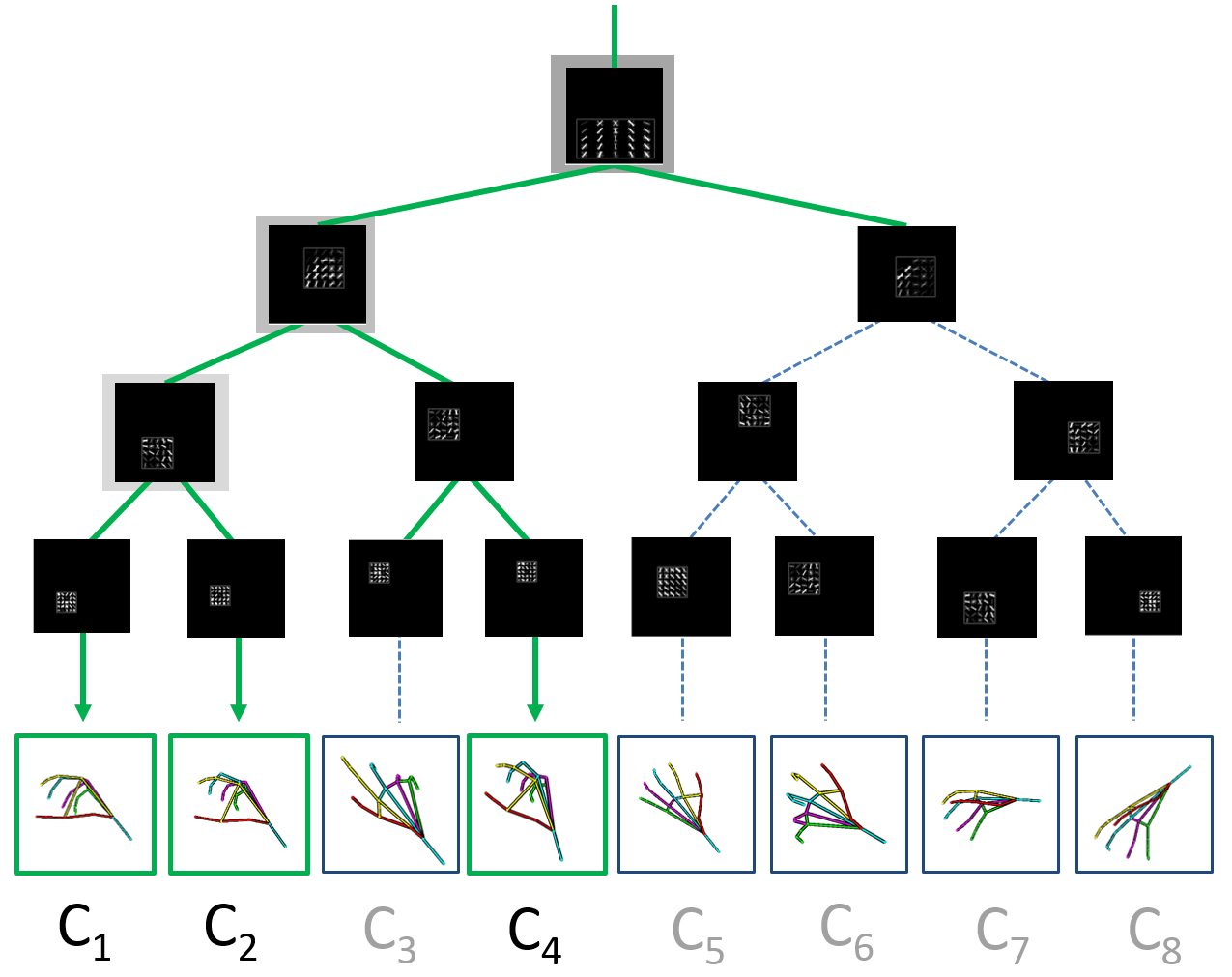}\\
\end{tabular}
\caption{{\bf Hierarchical cascades}. We approach pose-estimation as a $K$-way classification problem. We define a {\em linear-chain} cascade of rejectors for each of $K$ pose class (\textit{left}). By sharing ``weak'' classifiers across these cascades, we can efficiently organize the collection into a coarse-to-fine tree $G=(V,E)$ with $K$ leaves (\textit{right}). Weak classifiers near the root of tree are tuned to fire on a large collection of pose classes, while those near the leaves are specific to particular pose classes. Each linear cascade can be recovered from the tree by enumerating the ancestors of a leaf node.
  \label{fig:cascade}
}
\end{figure*}

{\bf Hierarchical quantization:} Firstly, we represent each training example as a depth image $x$ and a label vector $y$ of joint positions in a canonical coordinate frame with normalized position and scale. We quantize this space of poses $\{y\}$ into $K$ discrete values with $K$-mean clustering. We then agglomeratively merge these quantized poses into a hierarchical tree $G=(V,E)$ with $K$-leaves, following the procedure of \cite{RogezROT12}. Each node $i \in V$ represents a coarse pose class. We visualize the tree in Fig.~\ref{fig:hierarchy}. 

{\bf Coarse-to-fine sharing:} Given a test image $x$, a binary classifier tuned for coarse pose-class $i \in V$ is evaluated as: 
\begin{align}
f_i(x) = \prod_{j \in A_i} h_j(x) \quad \text{where} \quad h_j(x) = {\bf 1}_{[w_j^Tx > 0]} \label{eq:tree}
\end{align}
\noindent where ${\bf 1}$ is the indicator function which evaluates to 1 or 0. Here, $f_i$ is a ``strong'' classifier for class $i$ obtained by ANDing together ``weak'' binary predictions $h_j$ from a set $j \in A_i$. $A_i$ is the set of ``ancestor'' nodes encountered on the path from $i$ to the root of the tree ($i$, its parents, grandparents, etc.). Each weak classifier $h_j$ is a thresholded linear function that is defined on appearance features extracted from window $x$. We use HOG appearance features extracted from subregions within the window, meaning that $w_j$ can be interpreted as a zero-padded ``part'' template tuned for coarse-pose $j$. Parts higher in the tree tend to be generic and capture appearance features common to many pose classes. Parts lower in the tree, toward the leaves, tend to capture pose-specific details.

{\bf Breadth-first search (BFS):} The prediction for pose-class $i$ will be $1$ if and only if all classifiers in the ancestor set predict $1$. If node $i$ fails to fire, its children (and their descendants) cannot be detected and so can be immediately rejected. 
We can efficiently prune away large portions of the pose-space when evaluating region $x$ with a truncated breadth-first search (BFS) through graph $G$ (see Alg.~\ref{alg:classif1cascade}), making scanning-window evaluation at test-time quite efficient. Though a queue-based BFS is a natural implementation, we have not seen this explicitly described in previous work on hierarchical cascades~\cite{zehnder2008efficient,zehnder2008efficient,chenhierarchical,RogezROT12}. As we will show, such an ``algorithmic'' perspective immediately suggests straightforward improvements for training and modeling averaging.
\begin{algorithm}[htb]
\caption{Classification with a single cascade. We perform a truncated breadth-first search (BFS) using a first-in, first-out queue. We insert the children of the node $i$ into the queue only if the weak classifier $w_i$ successfully fires.
\label{alg:classif1cascade} 
}
\SetAlFnt{\tiny\sf}
\SetKwInOut{Input}{input} \SetKwInOut{Output}{output}

\Input{Image window $x$, classifiers $\{w_i, i \in V\}$}
\Output{$vote(i)$ for each leaf class $i$.}
\BlankLine 
create a queue Q\;
enqueue 1 onto Q\;
\While{$Q \neq \text{empty} $}{
i=Q.dequeue()\;
\If{$w_i^Tx >0$}{
\For {$ \forall k \in$ child(i)}
{
enqueue k onto Q\;
}
\If{$k=\emptyset$}
{vote(i)=1\;}
}
}
\end{algorithm}

{\bf Multi-class detection:} When Eq \eqref{eq:tree} is evaluated on leaf classes, it is mathematically equivalent to a collection of $K$ {\em linear chain} cascades tuned for particular poses. These linear chains are visualized in Fig.~\ref{fig:cascade}. From this perspective, multiple poses (or leaf classes) may fire on a single window $x$. This is in contrast to other hierarchical classifiers such as decision trees, where only a single leaf can be reached. We generally report the highest-scoring pose as the final result, but alternate high-scoring hypotheses may still be useful (since they can be later refined using say, a tracker). We define the score for a particular pose class by aggregating binary predictions over a large {\em ensemble} of cascades, as described below.

{\bf Ensembles of cascades:} To increase robustness, we aggregate predictions across an ensemble of classifiers. 
\cite{RogezROT12} describes an approach that makes use of a pool of weak part classifiers at node $i$:
\begin{align}
h_i(x) \in H_i \quad \text{where} \quad |H_i| = M
\end{align}
One can instantiate a tree by selecting a weak classifier (from its candidate pool $H_i$) for each node $i$ in graph $G=(V,E)$. This defines a potentially exponential-large set of instantiations $M^{|V|}$, where $M$ is the size of each candidate pool. In practice, \cite{RogezROT12} found that averaging predictions from a small random subset of trees significantly improved results.

\subsection{Joint training of exponential ensembles}

In this section, we present several improvements that apply in our problem domain. Because of local ambiguities due to self-occlusions, we expect individual part templates to be rather weak. This in turn may cause premature cascade rejections. We describe modifications that reduce premature rejections through joint training of weak classifiers and exponentially-large ensembles of cascades.

{\bf Sequential training:} Much previous work assumes pose-specific classifiers are given \cite{stenger2007estimating,chenhierarchical} or independently learned \cite{RogezROT12}. For example, \cite{RogezROT12} trains weak classifiers $w_i$ by treating all {\em all} training examples from pose-class/node $i$ as positives, and examples from {\em all} other poses as negatives. Instead, we use only the training examples that pass through the rejection cascade up to node $i$. This better reflects the scenario at test-time. This requires classifiers to be trained in a sequential fashion, in a similar coarse-to-fine BFS over nodes from the root to the leaves (Alg.~\ref{alg:seqtraining} ). 

\begin{algorithm}[htb]
\SetAlFnt{\tiny\sf}
\SetKwInOut{Input}{input} \SetKwInOut{Output}{output}

\Input{Training data $(x_n,y_n)$ and tree $G=(V,E)$.}
\Output{Weak part classifiers $\{w_i, i \in V\}$.}
\BlankLine 
create a queue Q\;
enqueue (1,$\{x_n\},\{y_n\}$) onto Q\;
\While{$Q \neq \text{empty}$}{
 (i,x,y)=Q.dequeue()\;
 $w_i$ =Train($x$,$y \in class_i$)\;
$(x,y) := \{ (x_n,y_n): w_i^T x_n >0\}$\;
\For {$ \forall k \in$ child(i)}
{
enqueue (k,x,y) onto Q\;
}
} 
\caption{Cascade sequential training. We perform a BFS through pose classes, training weak classifiers $w_i$ by enqueueing node indices $i$ and the training data $(x,y)$ that reaches that node. $Train$ returns a (linear SVM) model given training examples with binary labels. 
With a slight abuse of notation, $y \in class_i$ denotes a set of binary indicators that specify which examples belong to $class_i$, where $class_i$ is the set of leaf classes reachable through a BFS from node $i$.}
\label{alg:seqtraining} 
\end{algorithm}

{\bf Exponentially-large ensembles:} Rogez et al. \cite{RogezROT12} average votes across a small number (around hundred) of explicitly-constructed trees. By averaging over a larger set, we reduce the chance of a premature cascade rejection. We describe a simple procedure for exactly computing the average over the exponentially-large set of $M^{|V|}$ in Alg.~\ref{alg:classifExp}. Our insight is that one can compute an {\em implicit} summation (of votes) over the set by caching partial summations $t$ during the BFS. We refer the reader to the algorithm and caption for a detailed description. To train the pool of weak classifiers, we can leverage our sequential training procedure. Simply replace Lines 5 and 6 of Alg.~\ref{alg:seqtraining} with the following:
\begin{align}
  \{w_{ij}: j = 1\ldots M\} &= \text{TrainEnsemble}(x,y\in class_i)\\
  (x,y) &:= \{(x_n,y_n): \sum_j w_{ij}^T x_n > 0\}
\end{align}
\noindent where TrainEnsemble is a learning algorithm that returns an ensemble of $M$ models by randomly selecting subsets of training data or subsets of features. We select random subsets of features corresponding to local regions from window $x$. This allows the returned models $w_{ij}$ to be visualized as "parts" (Fig.~\ref{fig:cascade}).

\begin{algorithm}[h!]
\caption{Classification with exponentially large number of cascades. When processing a node $i$, all its associated weak classifiers $h_i \in H_i$ are evaluated. We keep track of the (exponentially-large) number of successful ensemble components by enqueing a running estimate $t$ and node index $i$. Once the queue is empty, $vote[i]$ is populated with the number of ensemble components that fired on leaf class $i$ (which is upper bounded by $M^{|V|}$).
\label{alg:classifExp} 
}
\SetAlFnt{\tiny\sf}
\SetKwInOut{Input}{input} \SetKwInOut{Output}{output}

\Input{Image window $x$, weak classifier pools $\{H_i, i \in V \}$}
\Output{ $vote(i)$ for each leaf class $i$.}
\BlankLine 
create a queue Q\;
enqueue (1,1) onto Q\;
\While{$Q \neq \text{empty} $}{
(i,t)=Q.dequeue()\;
$t:=t \cdot \sum_{h_i \in H_i} h_i(x)$\;
\If{$t>0$}{
\For {$ \forall k \in$ child(i)}
{
enqueue (k,t) onto Q\;
}
\If{$k=\emptyset$}
{vote(i)=t\;}
}
}
\end{algorithm}

\begin{figure}[htb]
\begin{tabular}{cc} 
  \hspace{-6mm} \footnotesize Detection rate & \footnotesize Processing Time \vspace{-1mm}\\
   \hspace{-6mm}\includegraphics[width=0.27\textwidth]{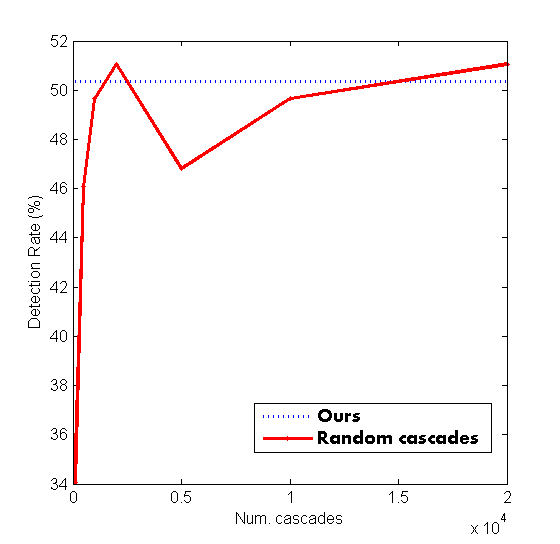} &
  \hspace{-6mm}\includegraphics[width=0.27\textwidth]{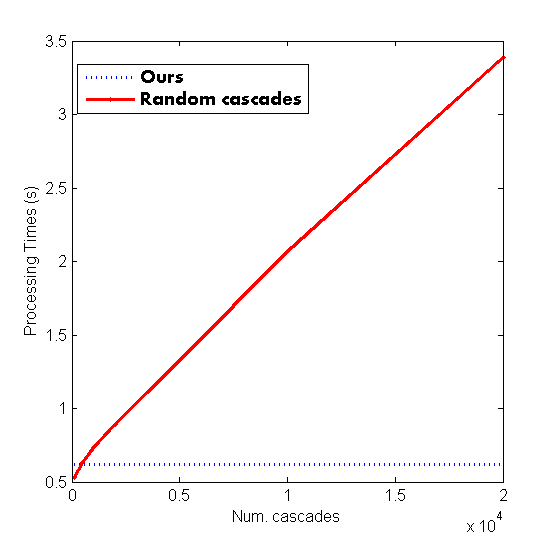} \\ (a)&(b) 
\end{tabular}
\caption{ {\bf Comparison with random cascades}. We show in (a) that our new detector is equivalent to an exponentially-large number of Random Cascades (RC) from \cite{RogezROT12}. In (b), we show that the RC computational cost increases linearly with the number of cascades and that, when considering a very large number of cascades, our model is more efficient. 
  \label{fig:Comparative-RC-Ensemble} 
}
\end{figure}

\begin{figure*}[htb]
  \centering
 \hspace{-0mm}\includegraphics[width=\textwidth]{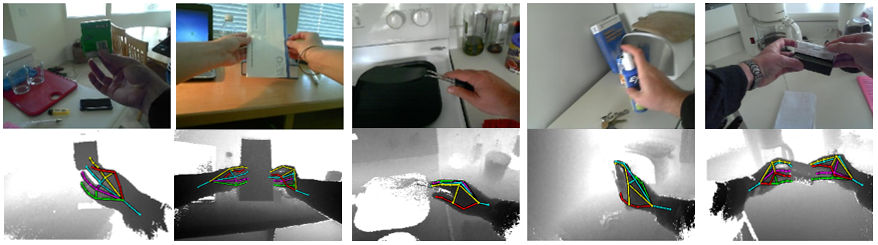} 
  \caption{\footnotesize {\bf Test data}.  We show  several examples of training RGB-D images captured with the chest-mounted Intel Creative camera from Fig.~\ref{fig:system}a. Real egocentric object manipulation scenes have been collected and annotated (full 3D hand poses) for evaluation.}
    \label{fig:testdataset}
\end{figure*}

\subsection{Implementation issues}
\label{sec:sparse}

{\bf Sparse search:} We leverage two additional assumptions to speed up our scanning-window cascades: 1) hands must lie in a valid range of depths, i.e., hands can not appear further away from the chest-mounted camera than physically possible and 2) hands tend to be of a canonical size $s$. These assumptions allow for a much sparser search compared to a classic scanning window, as only ``valid windows'' need be classified. A median filter is first applied to the depth map $d(x,y)$. Locations greater than arms length (75 cm) away are then pruned. Assuming a standard pinhole camera with focal length $f$, the expected image height of a hand at valid location $(x,y)$ is given by $S_{map}(x,y) = \frac{s}{f}d(x,y)$. We apply our hierarchical cascades to valid positions on a search grid (16-pixel strides in x-y direction) and quantized scales given by $S_{map}$, visualized as red dots in Fig.~\ref{fig:system}c.
 
{\bf Features}: We experiment with two additional sets of features $x$. Following much past work, we make use of HOG descriptors computed from RGB signals. We also evaluated oriented gradient histograms on depth images (HOG-D). While not as common, such a gradient-based depth descriptor can be shown to capture histograms of normal directions (since normals can be computed from the cross product of depth gradients) \cite{spinello2011people}. For depth, we use 5x5 HOG blocks and 16 signed orientation bins.  

\section{Experiments}
  \label{sec:exp}

{\bf Depth sensor:} Much recent work on depth-processing
has been driven by the consumer-grade PrimeSense sensor~\cite{sense2011primesensortmreference}, which is based on structured light technology. At its core, this approach relies on two-view
stereopsis (where correspondence estimation is made easier by active illumination). This may require large baselines between two views, which is undesirable for our egocentric application for two reasons; first, this requires larger form-factors, making the camera less mobile. Second, this produces
occlusions for points in the scene that are not visible in both views. Time-of-flight depth
sensing, while less popular, is based on a pulsed light emitter that can be placed arbitrarily close to the main camera, as no baseline is required. This produces smaller form factors
and reduces occlusions in that camera view. Specifically, we make use
of the consumer-grade TOF sensor from Creative \cite{Intel:PXC} (see Fig.~\ref{fig:system}a).


\hspace{15pt} {\bf Dataset:} We have collected and annotated (full 3D hand poses) our own benchmark dataset of real egocentric object manipulation scenes, which  we will release to spur further research \footnote{Please visit \url{www.gregrogez.net/}}. We developed a semi-automatic labelling tool which allows to accurately annotate partially occluded hands and fingers in 3D. A few 2D joints are first manually labelled in the image and used to select the closest synthetic exemplars in the training set. A full hand pose is then created combining the manual labelling and the selected 3D exemplar. This pose is manually refined, leading to the selection of a new exemplar, and the creation of a new pose. This iterative process is followed until an acceptable labelling is achieved.  We captured 4 sequences of 1000 frames each, which were annotated every 10 frames in both RGB and Depth. We use 2 different subjects (male/female) and 4 different indoor scenes. Some examples are presented in Fig.~\ref{fig:testdataset}. 

{\bf Parameters:} We train a cascade model trained with $K=100$ classes, a hierarchy of 6 levels and $M=3$ weak classifiers per node. We synthesize 100 training images per class. We experimented with larger numbers of classes (up to $K=1000$), but did not observe significant improvement. We suspect this is due to the restricted set of viewpoints and grasp poses present in egocentric interactions (which we see as a contribution of our work). As a point of contrast, we trained a non-egocentric hand baseline in Fig.~\ref{fig:Comparative-PR-3rdperson-Egocentric} that operated best at $K=800$ classes.

\subsection{Benchmark performance}
In this subsection, we validate our proposed architecture on a in-house dataset of 3rd-person hands. Our goal is to compare performance with standard baselines, verifying that our architecture is competitive. In this section, we make use of a generic set of 3rd-person views for training and use $K=800$ pose classes.
We then use this system as a starting point for egocentric analysis, exploring various configurations and priors further in the next section.
\label{sect:baselines}
\begin{figure*}[ht]
  \begin{tabular}{cccc}
    \centering
    \hspace{-6mm} \footnotesize 3rd-person hand detection & \hspace{-4mm}   \footnotesize 3rd-person finger tip detection   & \hspace{-6mm}   \footnotesize Egocentric hand detection &  \hspace{-4mm} \footnotesize Egocentric finger tip detection \vspace{-0mm}\\
    \hspace{-3mm}\includegraphics[trim=0.25cm 0.5cm 0.5cm 0.05cm, clip=true,width=0.25\textwidth]{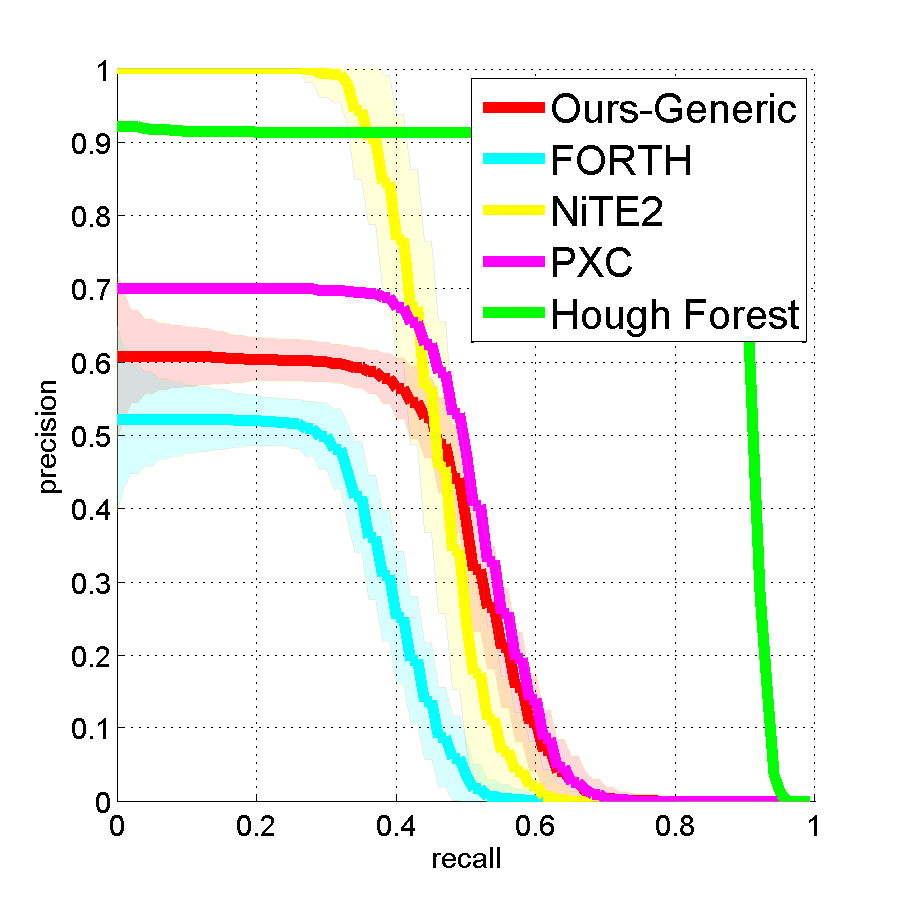} &
    \hspace{-3mm}\includegraphics[trim=0.25cm 0.5cm 0.5cm 0.05cm,clip=true,width=0.25\textwidth]{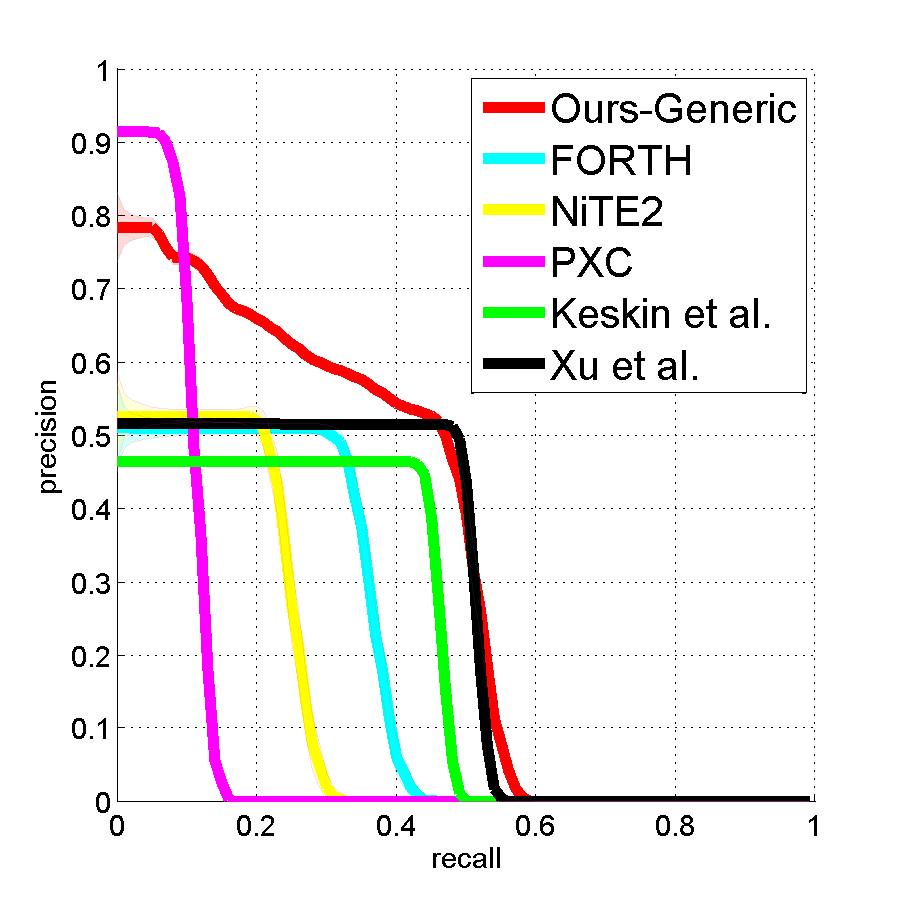} &
    \hspace{-3mm}\includegraphics[trim=0.25cm 0.5cm 0.5cm 0.05cm,clip=true,width=0.25\textwidth]{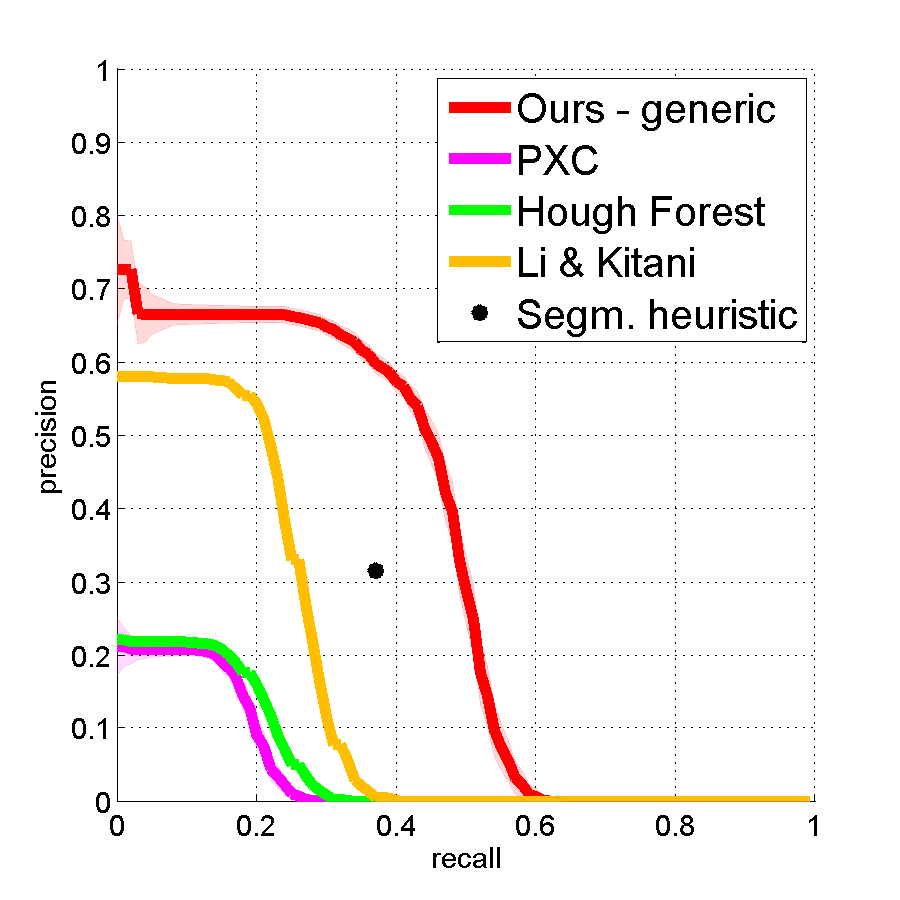} &
    \hspace{-3mm}\includegraphics[trim=0.25cm 0.5cm 0.5cm 0.05cm,clip=true,width=0.25\textwidth]{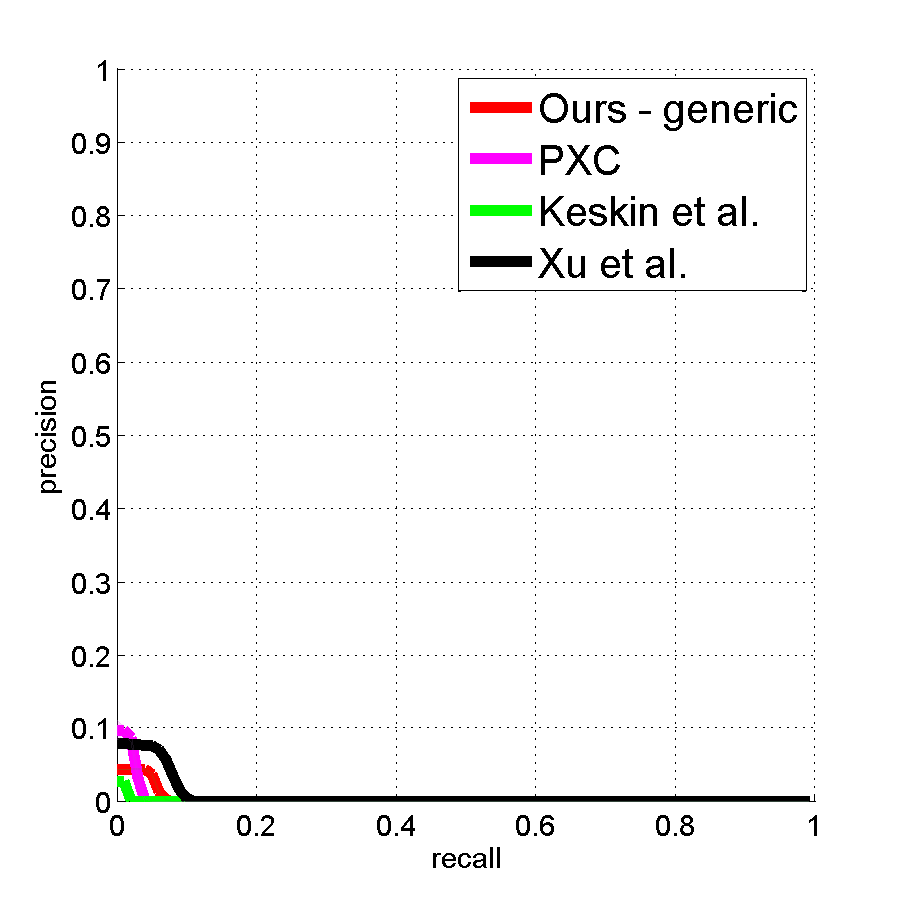} \\
    (a)&(b)&(c)&(d)
  \end{tabular}
  \caption{{\bf Third-person vs 1st person }. Numerical results  for 3rd-person (a-b) and egocentric (c-d) sequences. We compare our method (tuned for generic priors) to state-of-the-art techniques from industry (NITE2~\cite{NiTE2} and PXC~\cite{Intel:PXC}) and academia (FORTH ~\cite{bmvc2011oikonom}, Keskin et al ~\cite{keskin2012hand} and Xu et al. ~\cite{XuChe_iccv13}) in terms of (a) hand detection and (b) finger tips detection. 
  We refer the reader to the main text for additional description, but emphasize that (1) our method is competitive (or out-performs) prior art for detection and pose estimation and (2) pose estimation is considerably harder in egocentric views.
    \label{fig:Comparative-PR-3rdperson-Egocentric}
  }
\end{figure*}

{\bf Evaluation:} We evaluate both hand detections and pose estimation. A candidate {\em detection} is deemed correct if it sufficiently overlaps the ground-truth bounding-box (in terms of area of intersection over union) by at least 50\%. As some baseline systems report the pose of only confident fingers, we measure {\em finger-tip detection} accuracy as a proxy for pose estimation. To make this comparison fare, we only score visible finger tips and ignore occluded ones.

\textbf{Baselines: } We compare our method to
state-of-the-art techniques from industry~\cite{NiTE2,Intel:PXC} and
academia~\cite{bmvc2011oikonom,XuChe_iccv13,keskin2012hand,LiICCV13}. Because public code is not available, we re-implemented Xu et al\.~\cite{XuChe_iccv13} and Keskin et
al\.~\cite{keskin2012hand}, verifying that our performance matched published results in~\cite{TangCTK14}. Xu proposes a three stage
pipeline: (1) detect position and in-plane rotation with a Hough forest (2) estimate joint
angles with a second stage Hough forest (3) apply an articulated model
to validate global consistency of joint estimates. Keskin's model also
has three stages: (1) estimate global hand shape from local votes (2) given
the estimated shape, apply a shape-specific decision forest to predict a part label for each pixel
and (3) apply mean-shift to regress joint
positions. Because Keskin's model assumes detection is solved, we
experimented with several different first-stage detectors before
settling on using Xu's first stage Hough forest, due to its superior
performance. Thus, both models share the same hand detector in our evaluation. 

\begin{figure}[htb]
  \centering 
 Segmentation \hspace{20pt} Pose on depth \quad\hspace{20pt} RGB image \quad\quad\\
 \hspace{-0mm}\includegraphics[width=0.48\textwidth]{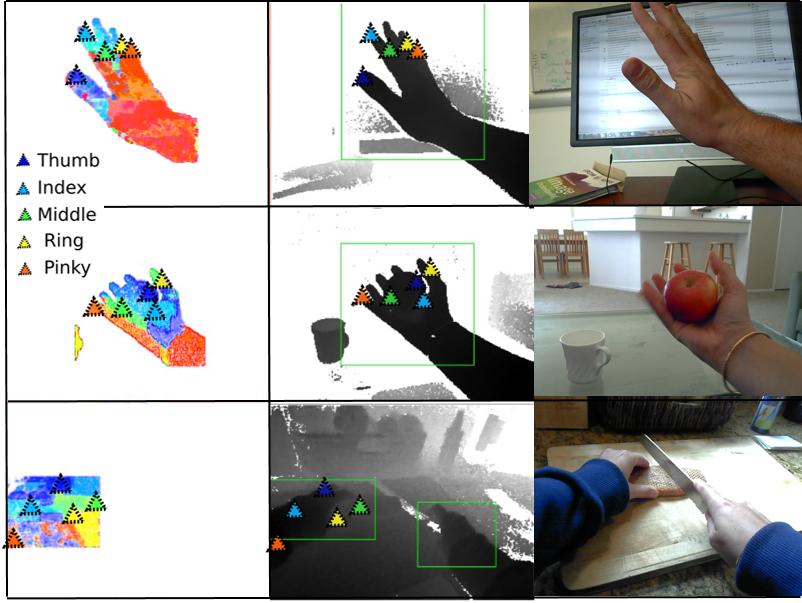} 
  \caption{\footnotesize {\bf Qualitative results obtained by state-of-the-art }\cite{keskin2012hand}: (top) Shows a test sample which is akin to those addressed by prior work. The hand is easily segmentable from the background and no object interaction is present. In this case, the method from \cite{keskin2012hand} correctly identifies 4 of the fingers and has a small localization error for the 5th.
(middle) Here the hand is holding a ball. Now, only the pinky is correctly localized, despite the algorithm being provided with object interaction in the training data. This demonstrates the combinatorial increase in difficultly posed by introducing objects.
(bottom) Finally, when we are able to correctly detect the  hand, but cannot easily segment it from the background, methods based on per-pixel classification fail because they produce strong ``garbage'' classifications for background.
}
    \label{fig:ego_pixel_classif}
\end{figure}

\textbf{Third-person vs egocentric:} Following Fig.~\ref{fig:Comparative-PR-3rdperson-Egocentric}, our hierarchical cascades are competitive for 3rd-person hand detection (a) and state-of-the-art for finger detection (b). When evaluating the same models (without retraining) on egocentric test data (c) and (d), most methods (including ours) perform significantly worse. FORTH and NITE2 trackers catastrophically fail since hands frequently leave the view, and so are omitted from (c) and (d).  Random Forest baselines \cite{XuChe_iccv13,keskin2012hand} drop in performance, even for hand detection. We posit this drop comes difficulties in segmenting egocentric hands (as shown in Fig.~\ref{fig:ego_pixel_classif}). To test this hypothesis, we develop a custom segmentation heuristic that looks for arm pixels near the image border, followed by connected-component segmentation. We also experiment with the (RGB) pixel-level hand detection algorithm from \cite{LiICCV13}. These segmentation algorithms outperform many baselines for hand-detection, but still underperforms our hierarchical cascade. We conclude that
(1) hand pose estimation is considerably harder in the egocentric
setting and (2) our (generic) pose estimation system is a
state-of-the-art starting point for our subsequent analysis. Finally, we posit that our strong performance (at least for hand detection) arises from our global hand classifiers, while most baselines tend to classify local parts.

\subsection{Diagnostic analysis}
\label{sect:diagnostic}

In this section, we further explore various configurations and priors for our approach, tuned for the egocentric setting.

{\bf Evaluation:} Since our algorithm always returns a full articulated hand pose, we  evaluate pose estimation with {\em 2D-RMS} re-projection error of keypoints. This time, we score the 20 keypoints defining the hand, including those which are occluded. We believe this is important, because numerous occlusions arise from egocentric viewpoints and object manipulation tasks. This evaluation criteria will give a better sense of how well our method actually recognizes global hand poses, even in case of partially occluded hands. For additional diagnosis, we categorize errors into detection failures, correct detections but incorrect viewpoint, and correct detection and viewpoint but incorrect articulated pose. Specifically, {\em viewpoint-consistent detections} are detections for which the RMS error of all 2D joint positions falls below a coarse threshold (10 pixels). {\em Conditional 2D RMS} error is the reprojection error for well-detected (viewpoint -consistent) hands. Finally, we also plot accuracy as a function of the number of {\em N candidate detections} per image. With enough hypotheses, accuracy must max out at 100\%, but we demonstrate that good accuracy is often achievable with a small number of candidates (which may later be re-ranked, by say, a tracker).
 \begin{figure*}{t!}
\begin{tabular}{cccc}
  \centering
  \hspace{-6mm} {\footnotesize VP Detection (PR)} & \hspace{-4mm} {\footnotesize 2D RMSE (N candidates)}   & \hspace{-6mm} \footnotesize VP Detection (N candidates) &  \hspace{-4mm} \footnotesize Cond 2D RMSE (N candidates) \vspace{-0mm}\\
  \hspace{-8mm}\includegraphics[trim=0.25cm 0.5cm 0.5cm 0.05cm, clip=true,width=0.29\textwidth]{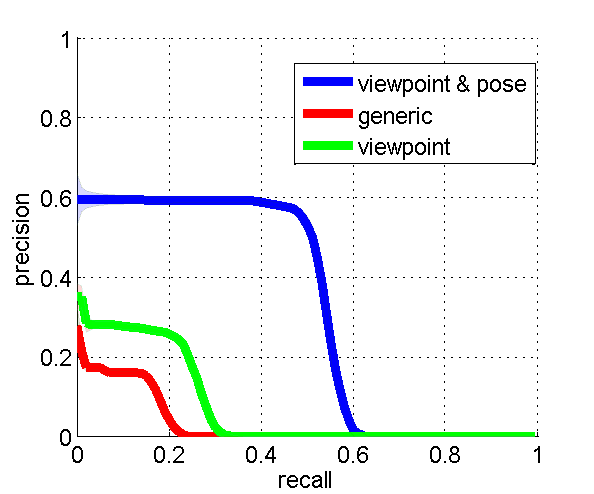} &
  \hspace{-4mm}\includegraphics[trim=0.25cm 0.5cm 0.5cm 0.05cm,clip=true,width=0.25\textwidth]{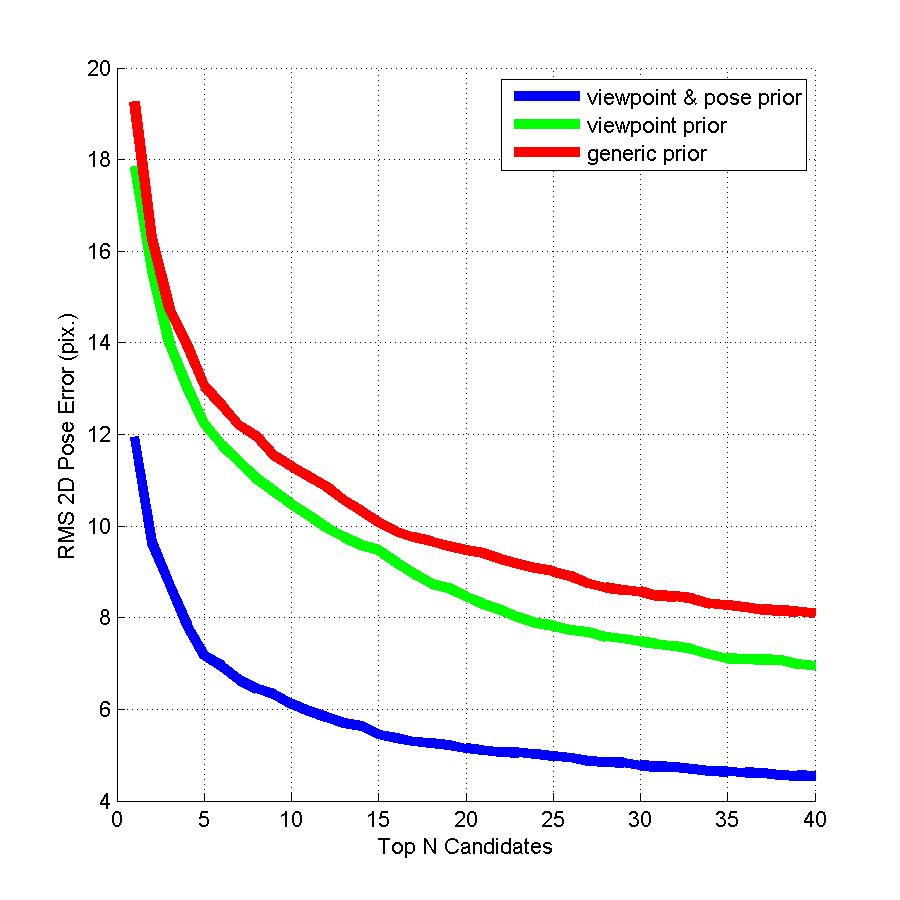} &
  \hspace{-4mm}\includegraphics[trim=0.25cm 0.5cm 0.5cm 0.05cm,clip=true,width=0.25\textwidth]{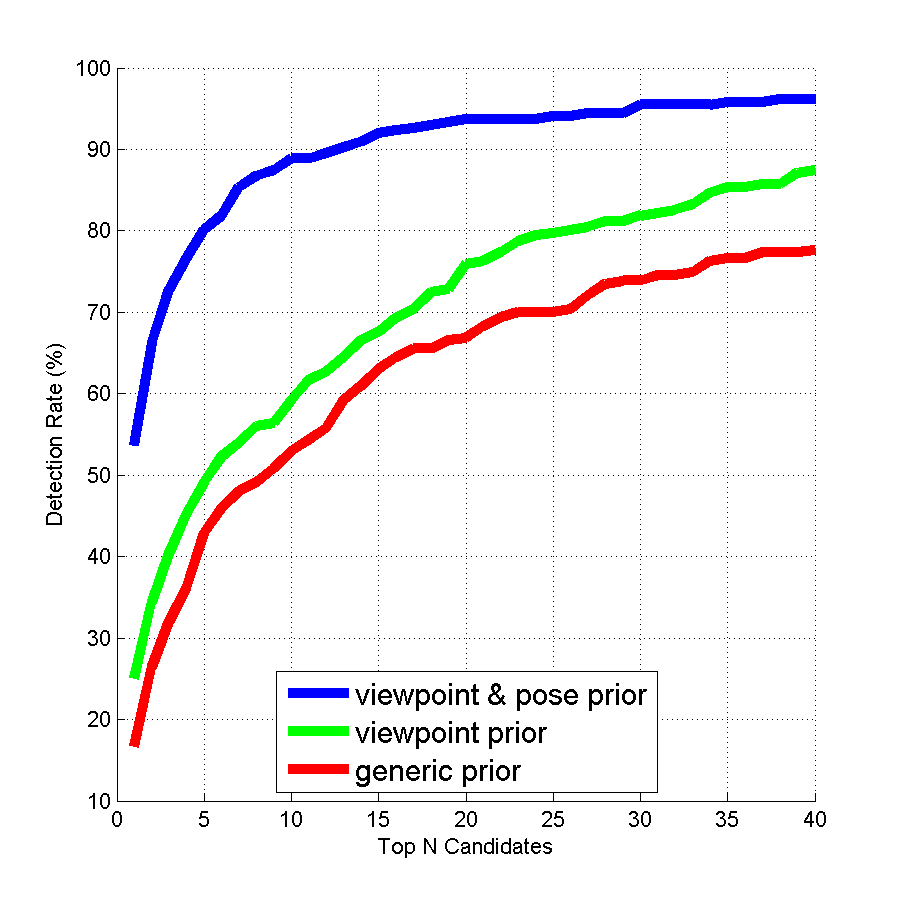} &
  \hspace{-4mm}\includegraphics[trim=0.25cm 0.5cm 0.5cm 0.05cm,clip=true,width=0.25\textwidth]{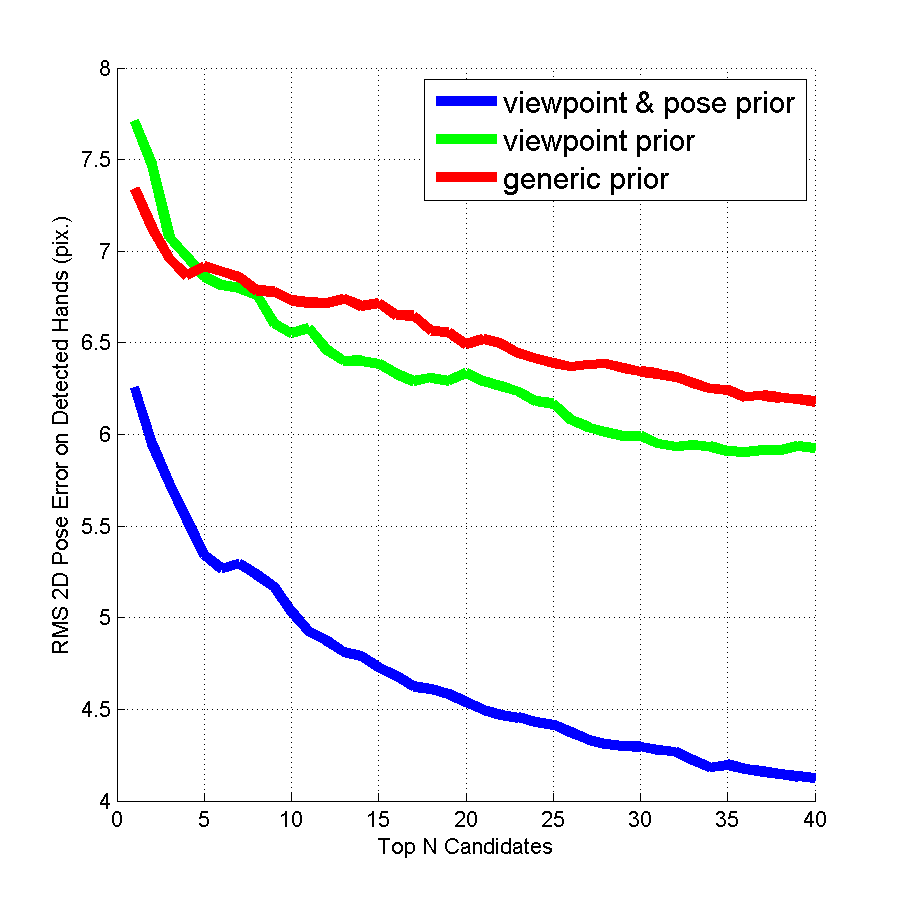} \\
  (a)&(b)&(c)&(d)
\end{tabular}
\caption{{\bf Quantitative results varying our prior}. We evaluate th different priors with respect to (a) viewpoint-consistent hand detection (precision-recall curve), (b) 2D RMS error, (c) viewpoint-consistent detections and (d) 2D RMS error conditioned on viewpoint-consistent detections. Please see text for detailed description of our evaluation criteria and analysis. In general, egocentric-pose priors considerably improve performance, validating our egocentric-synthesis engine from Sec.~\ref{sec:synth}. When tuned for $N=10$ candidates per image, our system produces pose hypotheses that appear accurate enough to initialize a tracker.
  \label{fig:Comparative-Pose-viewpoint-prior}
}
\end{figure*}

{\bf Pose+viewpoint prior:} We explore 3 different priors: 1) a \textit{generic prior}  obtained using a floating``libhand'' hand with all possible random camera viewpoints and pose configurations, 2) a \textit{viewpoint prior} obtained limiting a floating hand to valid egocentric viewpoints and 3) a \textit{ viewpoint \& pose prior} obtained using our full synthesis engine described in Sec.~\ref{sec:synth}, i.e. using a virtual egocentric camera mounted on a full body avatar manipulating objects. Note that we respectively consider 800, 140 and 100 classes to train these models. In Fig.~\ref{fig:Comparative-Pose-viewpoint-prior}, we show that, in general, a viewpoint prior produces a marginal improvement, while our full egocentric-specific pose and viewpoint prior considerably improves accuracy in all cases. This suggests that our synthesis algorithm correctly operationalizes egocentric viewpoint and pose priors, which in turn leads us to make better hypothesis for daily activities/grasp poses.
With a modest number of candidates $(N=10)$, our final system produces viewpoint-consistent detections in 90\% of the test frames with an average 2D RMS error of 5 pixels. From a qualitative perspective, this performance appears accurate enough to initialize a tracker. 
\begin{figure}[htb]
\begin{tabular}{cc}
  \centering 
   \footnotesize VP Detection (N candidates) &   \footnotesize Cond 2D RMSE (N candidates)\vspace{-0mm}\\
  \hspace{-5mm}\includegraphics[trim=0.25cm 0.5cm 0.5cm 1cm, clip=true,width=0.5\columnwidth]{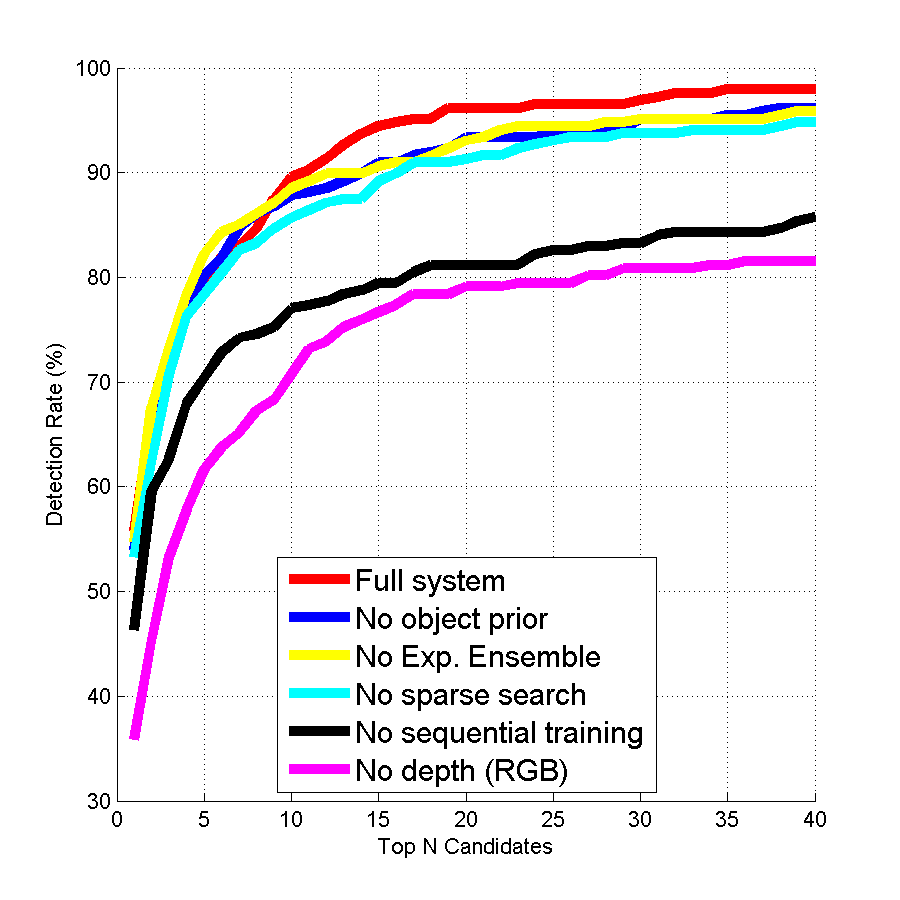} & 
  \includegraphics[trim=0.25cm 0.5cm 0.5cm 1cm,clip=true,width=0.5\columnwidth]{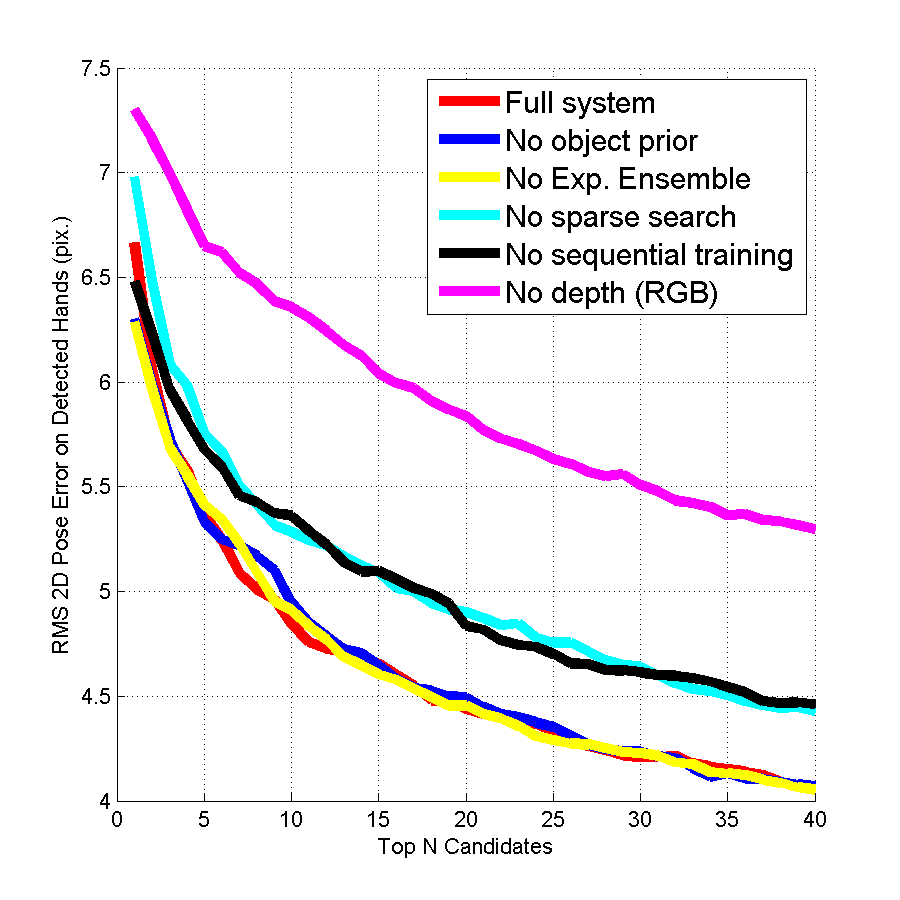} 
   \\ 
  (a)&(b)  
\end{tabular} 
\caption{{\bf Ablative analysis}. We evaluate performance when turning off particular aspects of out system, considering both (a) viewpoint-consistent detections (b) 2D RMS error conditioned on well-detected hands. When turning off our exponentially large ensemble or synthetic training, we use the default of 100-component ensemble as in~\cite{RogezROT12}. When turning off the depth feature, we use a classifier trained on aligned RGB images. Please see the text for further discussion of these results. 
  \label{fig:ablative}
}
\end{figure}
\begin{figure*}[htb]
\begin{tabular}{cccc} 
  \centering 
  \hspace{-6mm} {\footnotesize  VP Det. - Frames with Objects} & \hspace{-4mm} {\footnotesize Cond 2D RMSE - Frames with Obj.}   & \hspace{-6mm} \footnotesize VP Det. - Frames without Obj. &  \hspace{-4mm} \footnotesize Cond 2D RMSE - Frames without Obj. \vspace{-0mm}\\
  \hspace{-4mm}\includegraphics[trim=0.25cm 0.5cm 0.5cm 0.05cm, clip=true,width=0.25\textwidth]{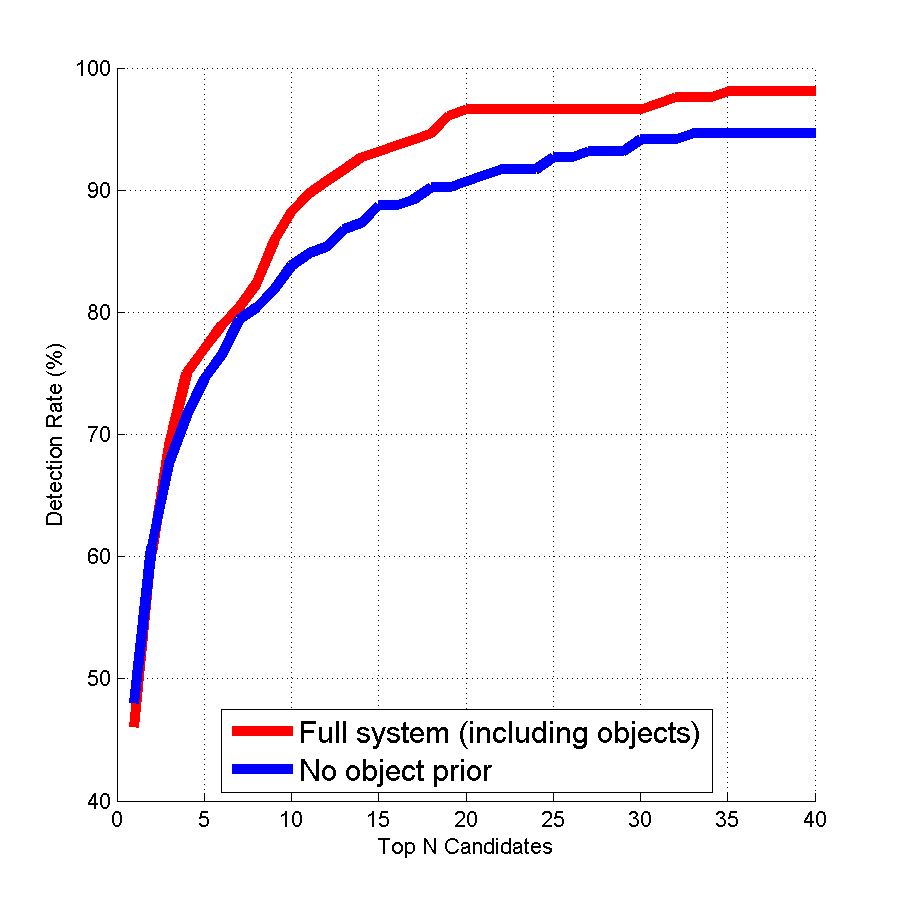} &
  \hspace{-3mm}\includegraphics[trim=0.25cm 0.5cm 0.5cm 0.05cm,clip=true,width=0.25\textwidth]{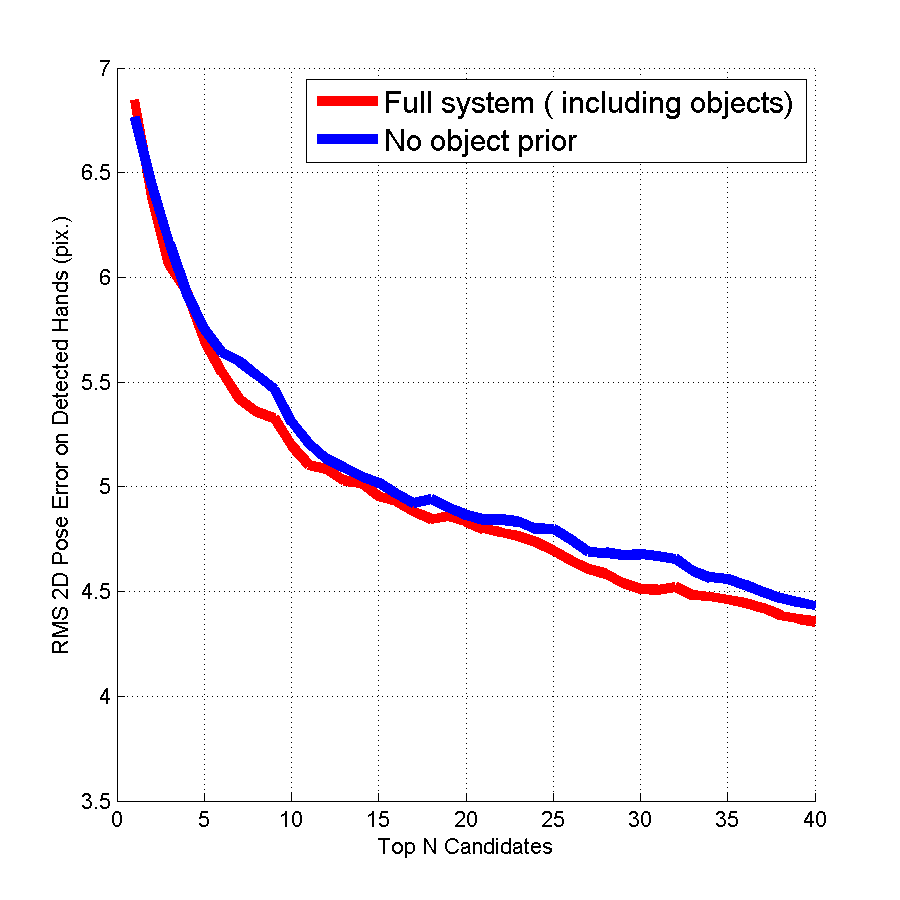} &
  \hspace{-3mm}\includegraphics[trim=0.25cm 0.5cm 0.5cm 0.05cm,clip=true,width=0.25\textwidth]{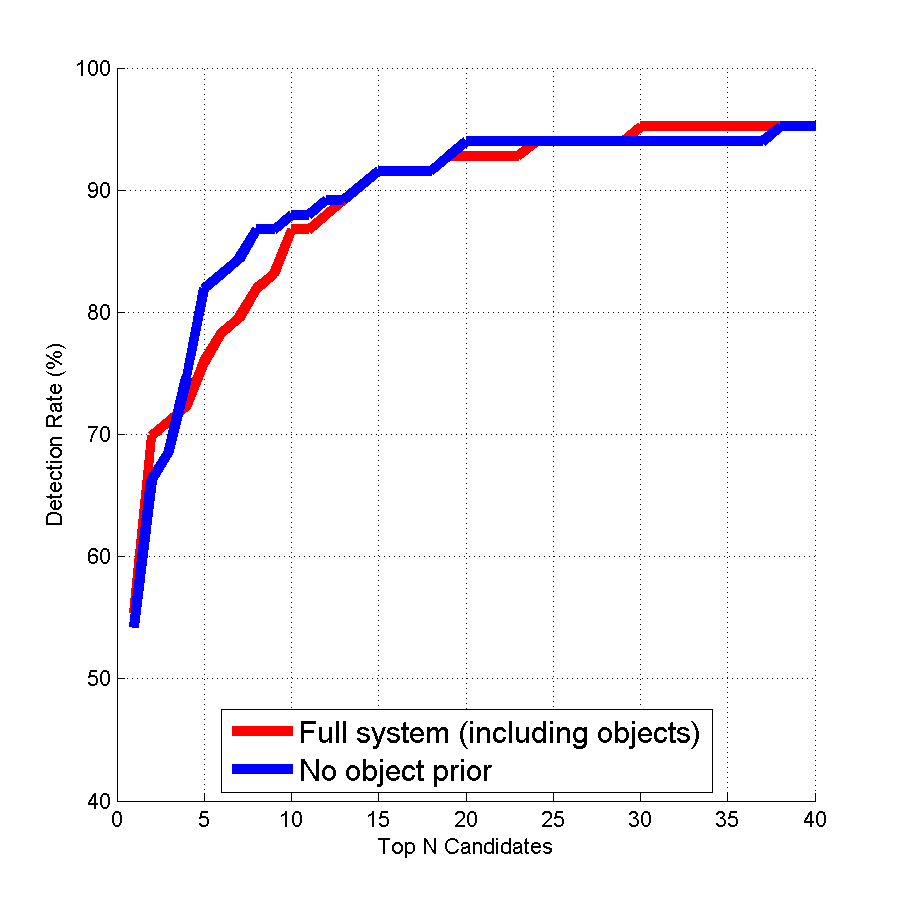} &
  \hspace{-2mm}\includegraphics[trim=0.25cm 0.5cm 0.5cm 0.05cm,clip=true,width=0.25\textwidth]{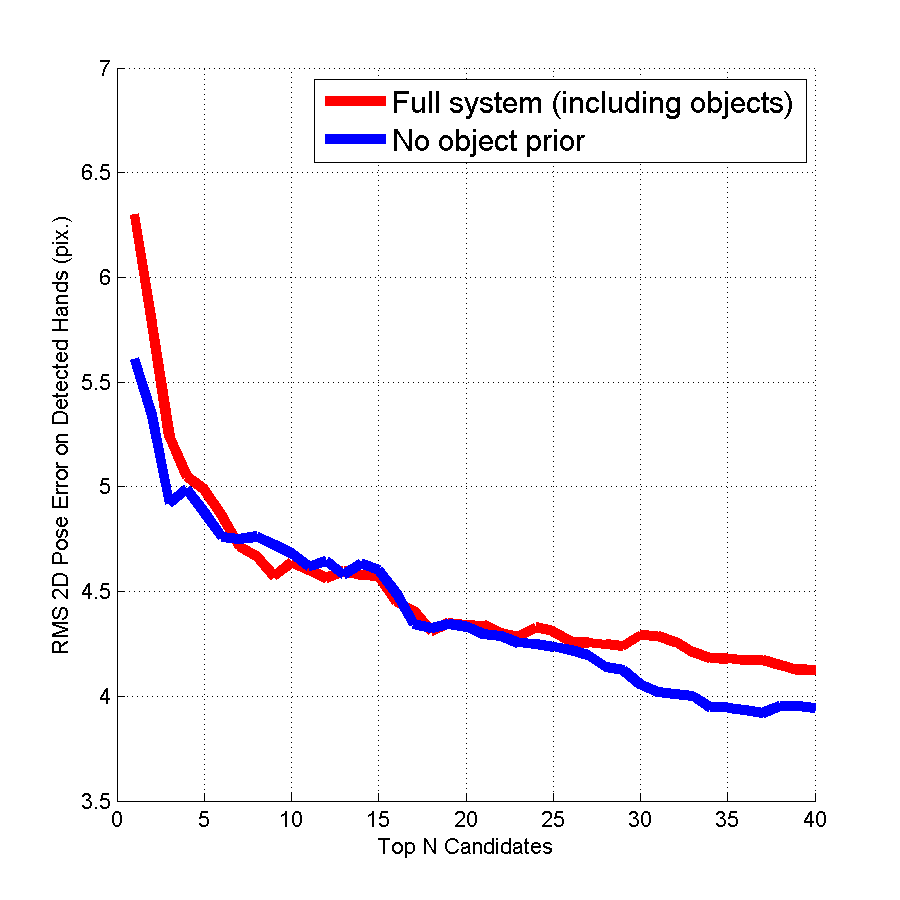} \\
  (a)&(b)&(c)&(d)
\end{tabular} 
\caption{{\bf Object prior }. Effect of modeling a hand with and without an object on hand detection (a and c) and hand pose recognition (b and d). Results are given for test frames with (a and b) and
  without (c and d)  object manipulation . Again, we measure viewpoint-consistent detections (a and c) and 2D RMS error conditioned on well-detected hands (b and d). Both hand detection and pose recognition are considerably more challenging for frames with object  interactions, likely due to additional occlusions from the manipulated objects. The use of an object prior provides a small but noticeable improvement for frames with objects,  but does not affect the performance of the system for the frames without objects.
 \label{fig:Comparative-WithWithout-Objects}
}
\end{figure*}

{\bf Ablative analysis:} To further analyze our system, we perform an ablative analysis that turns ``off'' different aspects of our system: sequential training, ensemble of cascades, depth feature, sparse search and additional object prior. Hand detection and conditional 2D hand RMS error are given  in Fig.~\ref{fig:ablative}. Depth HOG features and sequential training of parts are by far the crucial components of our system. Turning these parameters off decreases the detection rate by a substantial amount (between 10 and 30\%). Our exponentially-large ensemble of cascades and sparse search marginally improve accuracy but are much more efficient: in average, the exponentially-large ensemble is 2.5 times faster than an explicit search over a 100-element ensemble (as in \cite{RogezROT12}),  while the sparse search is 3.15 times faster than a dense grid. 
Modeling objects produces better detections, particularly for larger numbers of candidates. In general, we find this additional prior helps more for those test frames with object manipulations as detailed below.


{\bf Modeling objects:} 
In Fig.~\ref{fig:Comparative-WithWithout-Objects}, we analyze the effect of object interactions on egocentric hand detection and pose estimation when employing an object prior (or not). We plot the accuracy (for both viewpoint-consistent detections and conditional 2D RMS error) on those test frames with (a,b) and without (c,d) object interactions. The corresponding plots computed on the whole dataset (using both types of frames) is already shown in Fig.~\ref{fig:ablative}a and Fig.~\ref{fig:ablative}b . We see that additional modeling of interacting hands and objects (with an object prior) somewhat improves performance for frames with object manipulation without affecting the performance of the system for the frames without objects.

\begin{figure}[htb]
\begin{tabular}{cc}
  \centering 
  \footnotesize VP Detection (N candidates) &   \footnotesize Cond 2D RMSE (N candidates)\\
  \hspace{-5mm}\includegraphics[trim=0.25cm 0.5cm 0.5cm 1cm, clip=true,width=0.5\columnwidth]{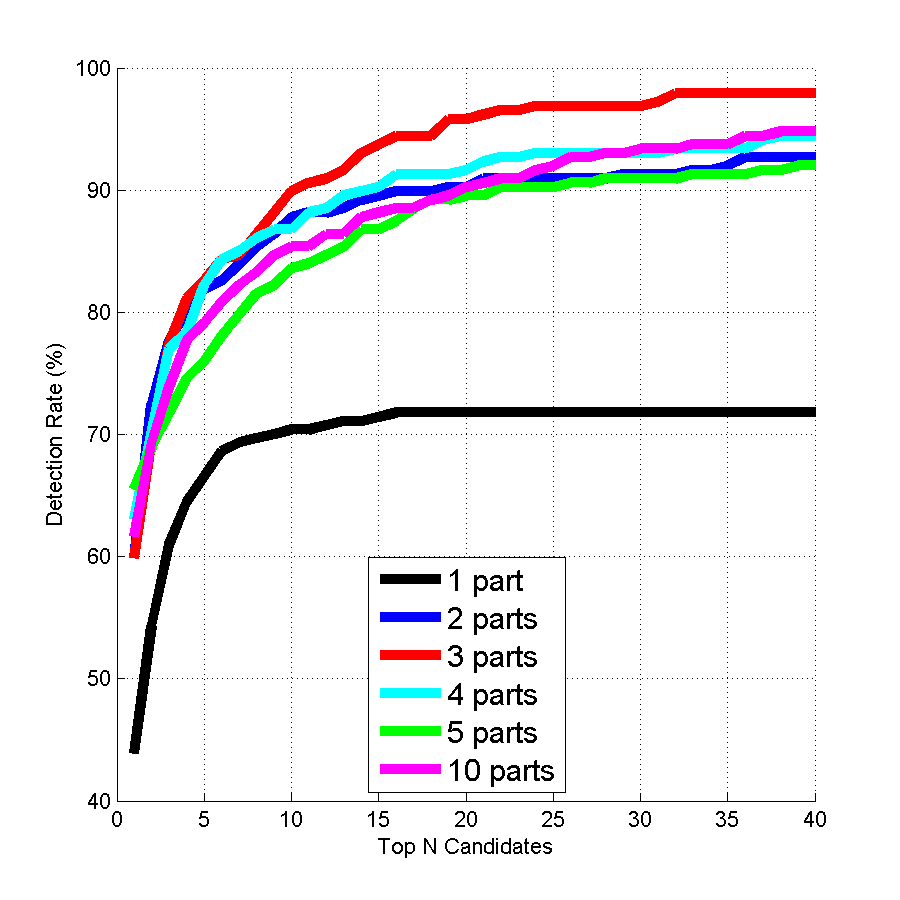} &
  \includegraphics[trim=0.25cm 0.5cm 0.5cm 1cm,clip=true,width=0.5\columnwidth]{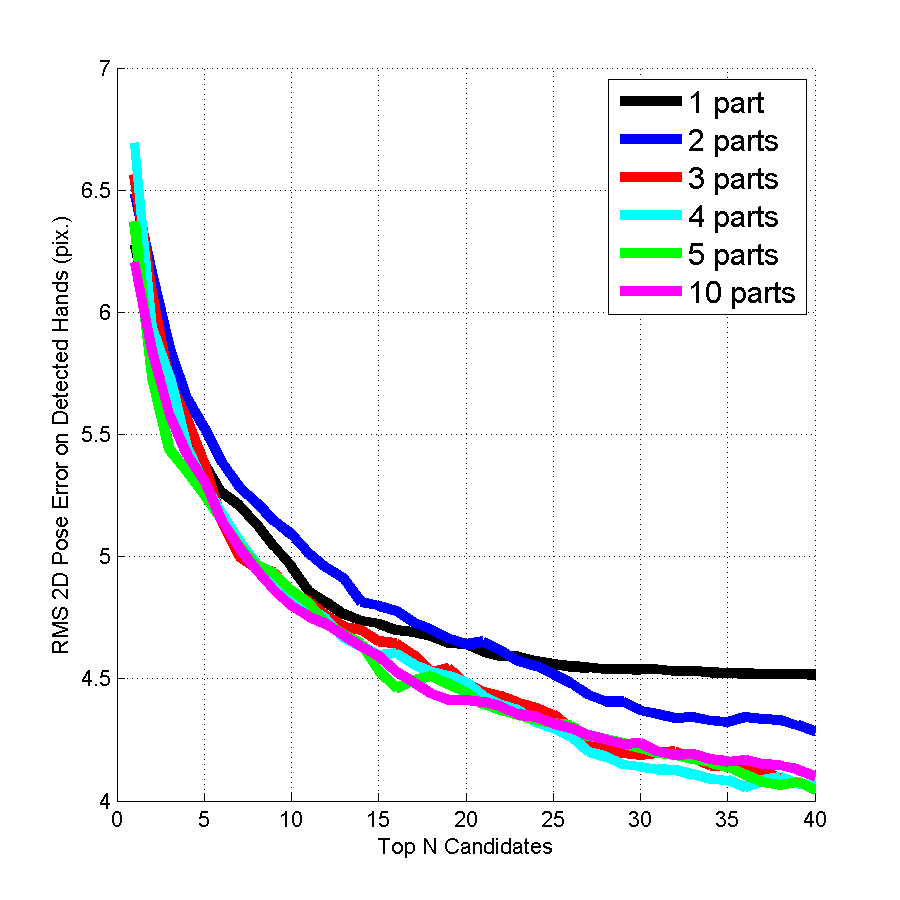} 
   \\ 
  (a)&(b)  
\end{tabular} 
\caption{ {\bf Choice of the number of parts}. We evaluate the importance of choosing the right number of parts in our cascade model by analyzing the performance achieved for different numbers. For each case, we compute the viewpoint-consistent detections (a) and 2D RMS error conditioned on well-detected hands (b) when varying the number of possible candidates. 
  \label{fig:numParts}
} 
\end{figure}

{\bf Number of parts:} In Fig.~\ref{fig:numParts}, we show the effect of varying the number of parts $M$ at each branch of our cascade model. We analyze both hand detection rate and hand pose average accuracy. These plots clearly validate our choice of using $M=3$ parts per branch  (Fig.~\ref{fig:numParts}a). The performance decreases when considering more parts because the classifier is more likely to produce a larger amount of false positives. Additionally, we can see in Fig.~\ref{fig:numParts}b that using more than 3 parts does not improve the accuracy in terms of hand pose.

\begin{figure*}[htb]
\begin{tabular}{cccc} 
  \centering
  \hspace{-6mm} {\footnotesize VP Detection (N candidates) } & \hspace{-4mm} {\footnotesize Cond 2D RMSE (N candidates) }   & \hspace{-6mm} {\footnotesize Hand Detection (10-pixel threshold)}   &  \hspace{-4mm} {\footnotesize Hand Detection (5-pixel threshold)}   \vspace{-0mm}\\
  \hspace{-4mm}\includegraphics[trim=0.25cm 0.5cm 0.5cm 0.05cm, clip=true,width=0.25\textwidth]{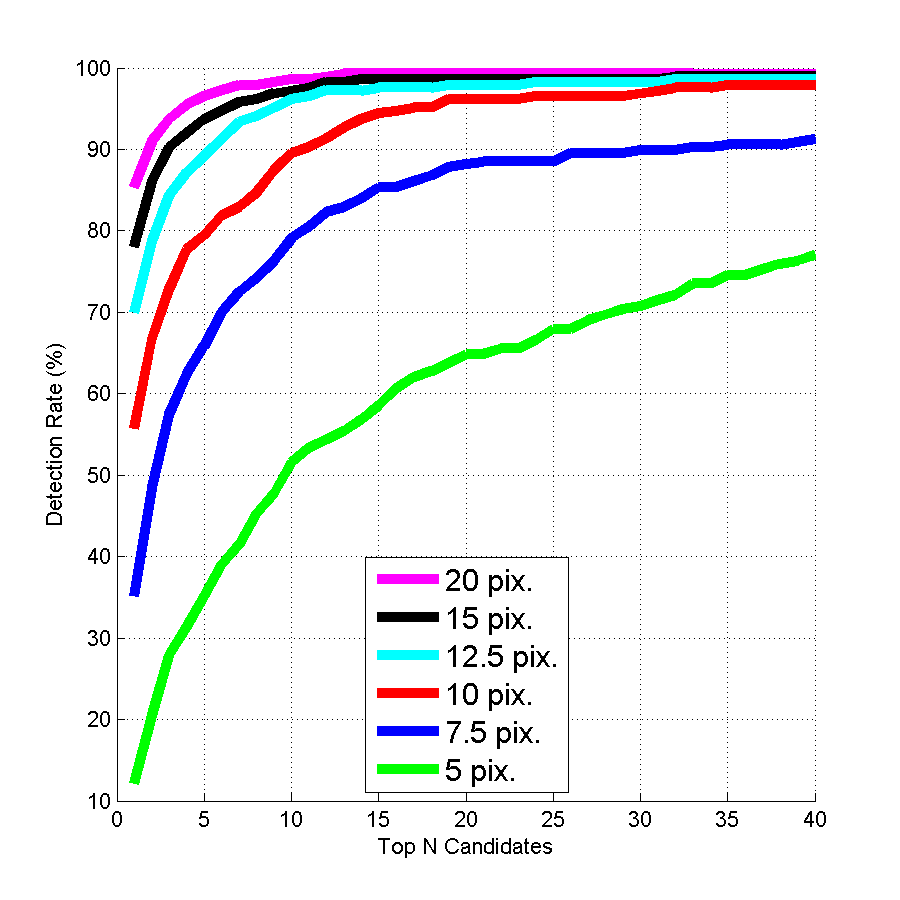} &
  \hspace{-3mm}\includegraphics[trim=0.25cm 0.5cm 0.5cm 0.05cm,clip=true,width=0.25\textwidth]{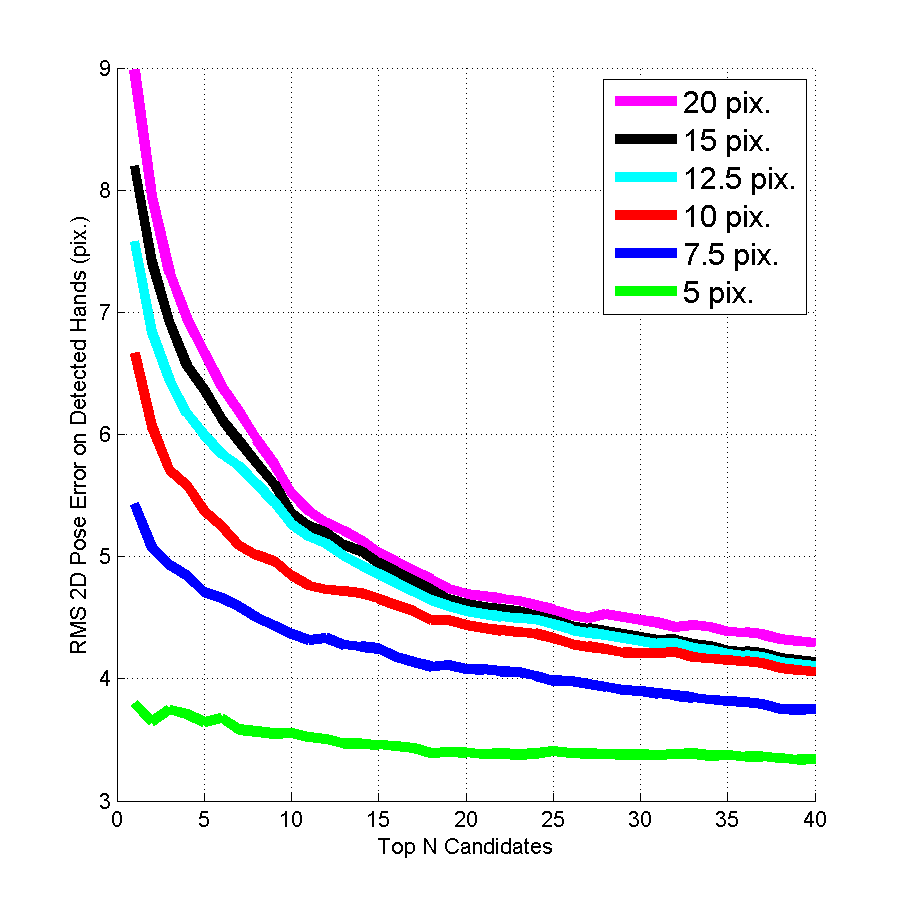} &
  \hspace{-3mm}\includegraphics[trim=0.25cm 0.5cm 0.5cm 0.05cm,clip=true,width=0.25\textwidth]{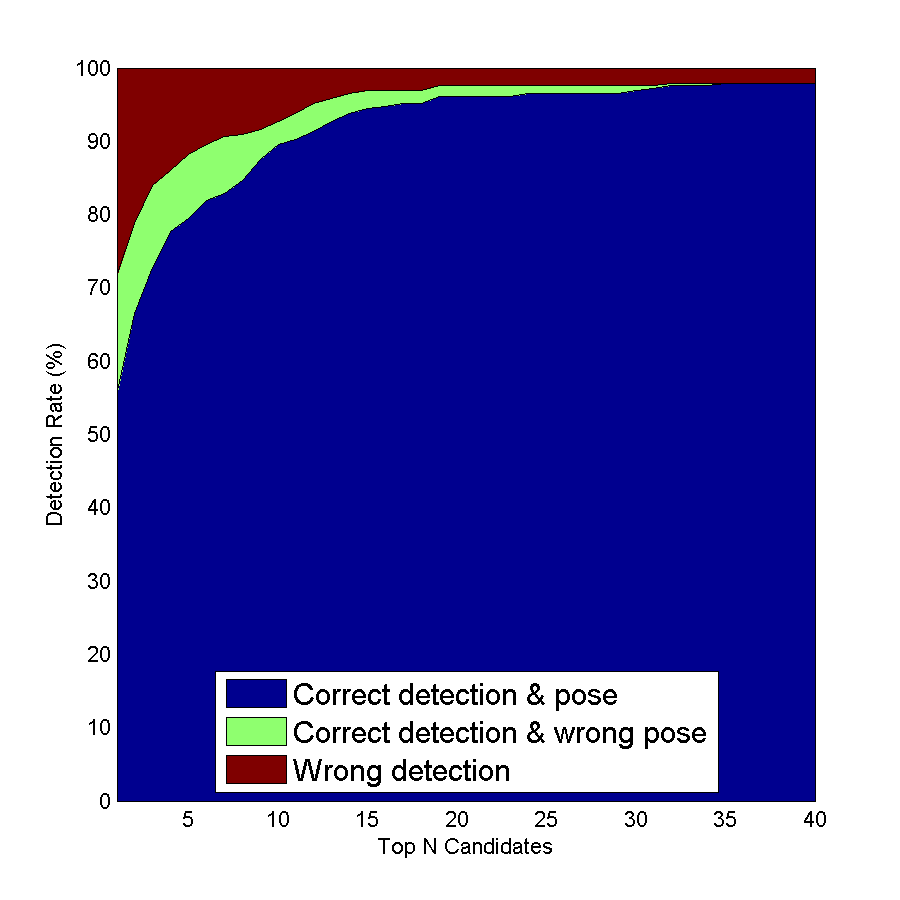} &
  \hspace{-2mm}\includegraphics[trim=0.25cm 0.5cm 0.5cm 0.05cm,clip=true,width=0.25\textwidth]{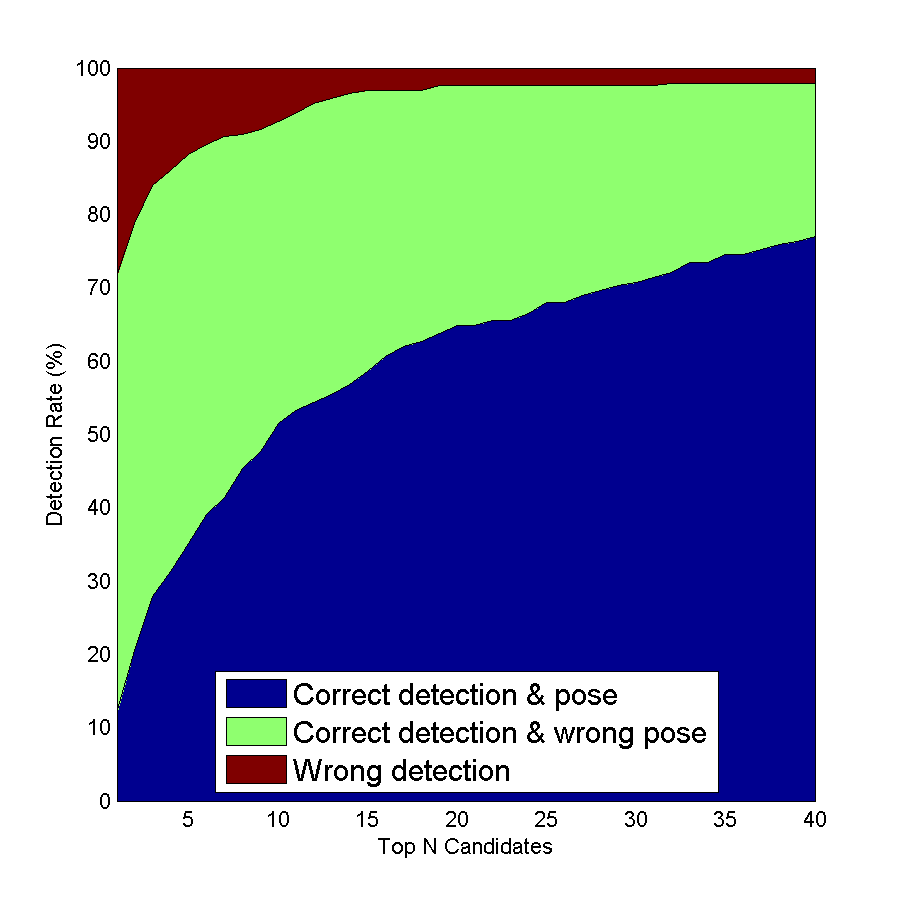}   \\
  (a)&(b)&(c)&(d)
\end{tabular} 
\caption{ {\bf Pixel threshold}. Effect of the pixel threshold on viewpoint-consistent detections (a) and 2D RMS error conditioned on well-detected hands(b). A lower pixel threshold decreases detection rate but increases pose accuracy on detected hands, while a higher threshold increases detection rate and decreases pose accuracy. We chose to use a threshold of 10 pixels as trade-off between these 2 criteria. In c and d, we show the percentage of detection and pose correctness for 2 different thresholds: 10 pixels (c) used all over the main paper and 5 pixels (d). The green area corresponds to correct detections (in term of bounding boxes overlap between estimated and ground-truth poses) but incorrect poses. The measure we use for detection (blue area) is much more strict as it only considers valid a detection when the hand pose is also correctly estimated. 
  \label{fig:Comparative-threshold}
}
\end{figure*}

{\bf Pixel  threshold:} In Fig.~\ref{fig:Comparative-threshold}, we show the effect of varying the pixel-overlap threshold for computing correct detection and average pose accuracy. A lower pixel threshold decreases detection rate but increases pose accuracy on detected hands and vice-versa. Our 10-pixel threshold is a good trade-off between these 2 criteria. In (c) and (d), we show the percentage of detection and pose correctness for 2 different thresholds: 10 pixels (c) used throughout the main paper and 5 pixels (d). The measure we used for detection (blue area) is much more strict than a simple bounding boxes overlap criteria (green area) as it only considers valid detections when the hand pose is also correctly estimated.
  
\begin{figure*}[t!]
	\includegraphics[width=\textwidth]{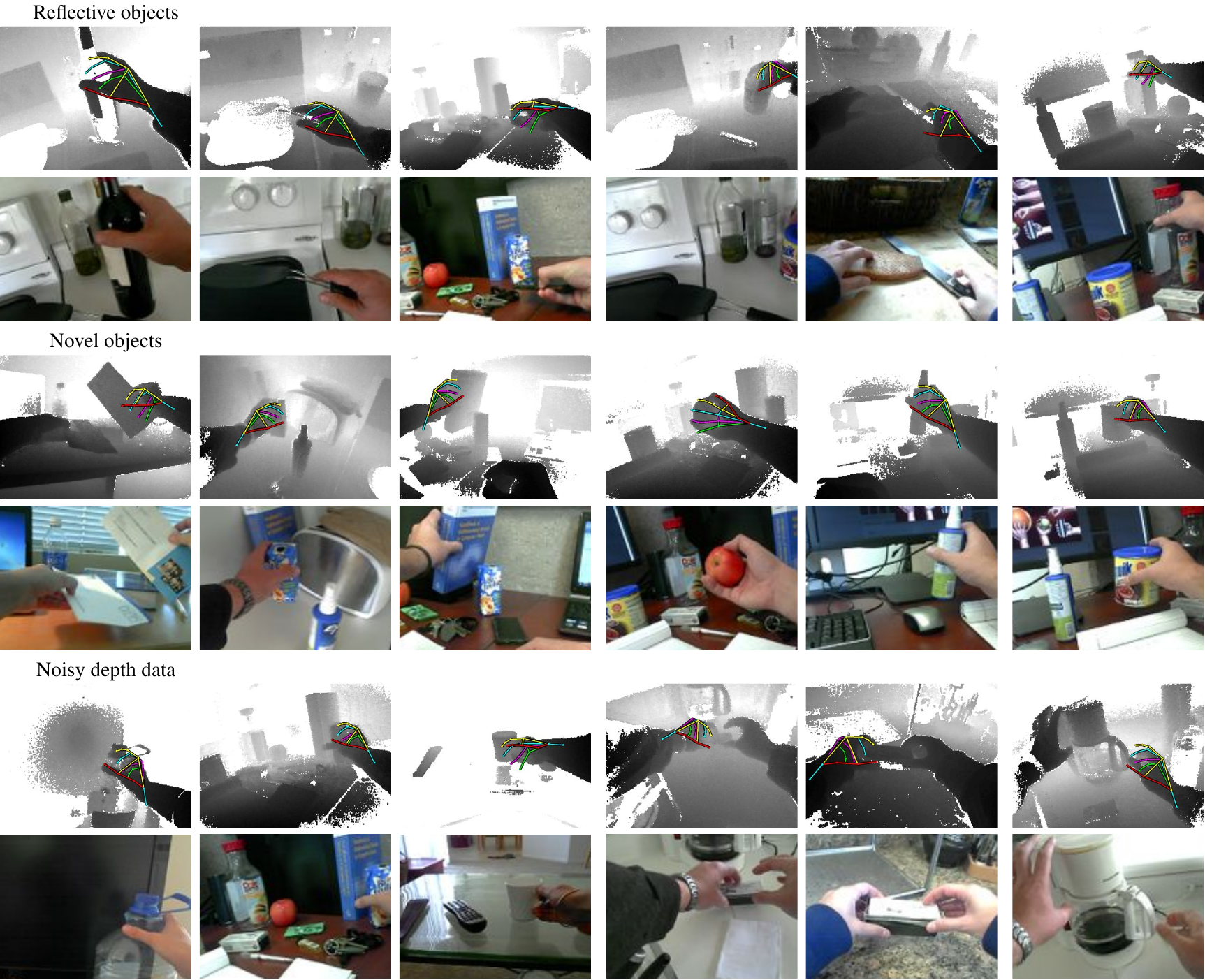}
\caption{{\bf Good detections}. We show a sample of challenging frames where the hand is correctly detected by our system. Reflective objects ({\bf top row:} wine bottle, pan, phone, knife and  plastic bottle) produce incorrect depth maps due to interactions with our sensor's infrared illuminant. Novel objects ({\bf middle row:} envelope, juice box, book, apple, spray and chocolate powder box) require generalization to objects not synthesized at train-time, while noisy depth data ({\bf bottom row}) showcases the robustness of our system.  
  \label{fig:easycases}
}
\end{figure*}

\begin{figure*}[htb]
  \centering
  \includegraphics[width=\textwidth]{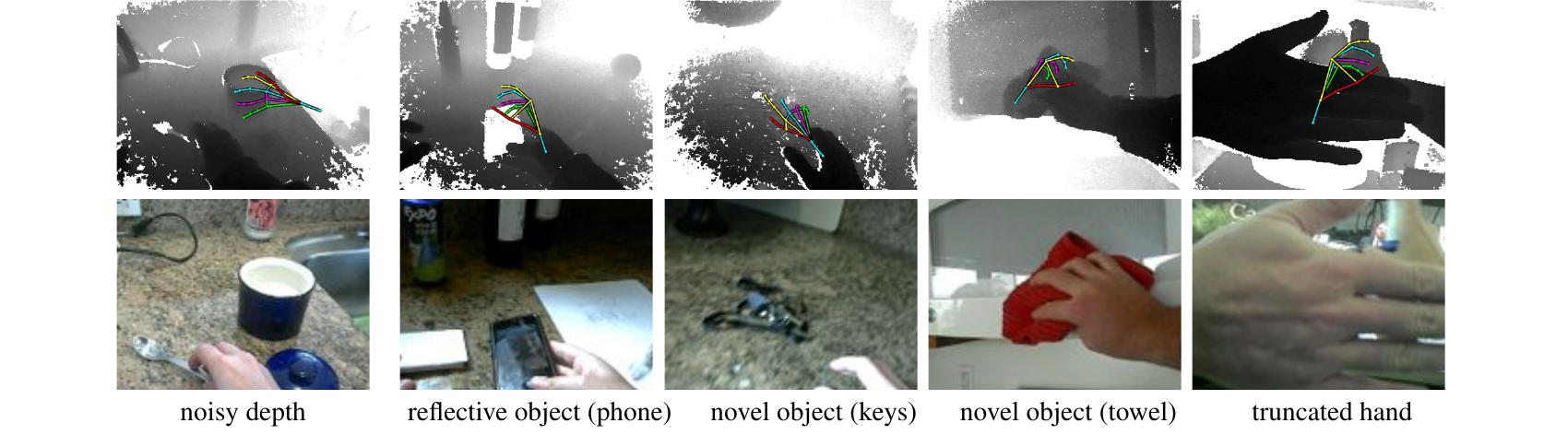}
\caption{{\bf Hard cases}. We show frames where the hand is not correctly detected by our system, even with 40 candidates. These hard cases include excessively-noisy depth data, hands manipulating reflective material (phone) or unseen/deformable objects that look considerably different from those in our training set (e.g. keys, towels), and truncated hands.
  \label{fig:hardcases}
}
\end{figure*}

{\bf Qualitative results:} We invite the reader to view our supplementary videos. We illustrate successes in difficult scenarios in Fig.~\ref{fig:easycases} and analyze common failure modes in Fig.~\ref{fig:hardcases}. Please see the figures for additional discussion.

\section{Conclusion}

We have focused on the task of hand pose estimation from egocentric viewpoints. For this problem specification, we have shown that TOF depth sensors are particularly informative for extracting near-field interactions of the camera wearer with his/her environment. We describe a detailed, computer graphics model for generating egocentric training data with realistic full-body and object interactions. We use this data to train discriminative K-way classifiers for quantized pose estimation. To deal with a large number of classes, we advance previous methods for hierarchical cascades of multi-class rejectors, both in terms of accuracy and speed.
Finally, we have provided an insightful analysis of the performance of our algorithm on a new real-world annotated dataset of egocentric scenes. Our method provides state-of-the-art performance for both hand detection and pose estimation in egocentric RGB-D images.

\bibliographystyle{spmpsci}
\bibliography{deva_bib}

\end{document}